\documentclass[letterpaper, 10 pt, conference]{ieeeconf}  

\IEEEoverridecommandlockouts                              

\usepackage{mathtools}
\usepackage{bm} 
\usepackage{amssymb}
\usepackage{amsfonts}
\usepackage{mathbbol} 
\usepackage{graphicx, import} 
\usepackage{float}
\usepackage{times}
\usepackage{multicol}
\usepackage{multirow}    
\usepackage{microtype}
\usepackage[all]{nowidow}
\usepackage{cases}
\usepackage{yfonts}
\usepackage{subcaption}
\usepackage{colortbl}
\usepackage{xparse}
\usepackage{arydshln}
\usepackage{microtype}
\usepackage{soul}
\usepackage[mediumspace,mediumqspace,squaren]{SIunits}

\usepackage{tabularx,booktabs}

\title{\LARGE \bf
Optimization of Humanoid Robot Designs for \\ Human-Robot Ergonomic Payload Lifting}
\author{Carlotta Sartore$^{1,2}$, Lorenzo Rapetti$^{1,2}$, and Daniele Pucci$^{1,2}$
\thanks{*The paper was supported by the Italian National Institute for Insurance against Accidents at Work (INAIL) ergoCub Project.}
\thanks{$^{1}$Artificial and Mechanical Intelligence  at Istituto Italiano di Tecnologia,
Center for Robotics Technologies, Genova, Italy.
        {\tt\small name.surname@iit.it}}%
\thanks{$^{2}$Machine Learning and Optimisation, The University of Manchester,
Manchester, United Kingdom.}%
}
\begin{document}
\maketitle
\thispagestyle{empty}
\pagestyle{empty}

\begin{abstract}
When a human and a humanoid robot collaborate physically,
\emph{ergonomics} is a key factor to consider. Assuming a given humanoid robot, several control architectures exist nowadays to address ergonomic physical human-robot collaboration.
This paper takes one step further by considering robot hardware parameters as optimization variables in the problem of collaborative payload lifting.
The variables that parametrize robot's kinematics and dynamics ensure their physical consistency, and the human model is considered in the optimization problem.
By leveraging the proposed modelling framework, 
the ergonomy of the interaction is maximized, here given by the agents' energy expenditure. Robot kinematic, dynamics, hardware constraints and human geometries are considered when solving the associated optimization problem. 
The proposed methodology is used to identify optimum hardware parameters for the design of the ergoCub robot, a humanoid possessing a degree of embodied intelligence for ergonomic interaction with humans. For the optimization problem, the starting point is the iCub humanoid robot. 
The obtained robot design reaches loads at heights in the range of $0.8-1.5 \text{ } \meter$ with respect to the iCub robot whose range is limited to $0.8-1.2 \text{ }$ \meter. The robot energy expenditure is decreased by about $33\%$, meanwhile, the human ergonomy is preserved, leading overall to an improved interaction.
\end{abstract}

\section{INTRODUCTION}
\label{sec:introduction}
In recent years, there has been a drastic improvement in the robotic field concerning the development and control of humanoid robots with the final aim of introducing such platforms in our daily life, endowing them with the ability to interact with human beings performing a large variety of tasks \cite{yokoyama2003cooperative,romano2017codyco,agravante2019human}.
One of the foreseen scenarios is the physical collaboration between humans and robots, which attracted the attention of the scientific community a long time ago, with a particular focus on the introduction of cobots into the industry 4.0 \cite{robotics2020}. Human-centered and ergonomic design are well-known aspects in engineering \cite{rasmussen2002design} and, when a human interacts with a robot, agents safety, and system efficiency are two of the principal aspects to be considered \cite{alami2006}. Several works have addressed the design of control architectures to achieve an \textit{ergonomic} physical human-robot interaction \cite{koeppe2005robot,vysocky2016human}, but the humanoid robot hardware design has been seldom considered as an element to be optimized for the specific collaborative action, and it is rather assumed as given. This paper takes a step towards conceiving optimal humanoid robot designs 
to maximize the ergonomics of human-robot collaborative tasks.

In a broader view, the methods presented in this paper fall into the category of \emph{embodied cognition}, which
shows how the emergence of intelligent behaviors is obtained via a synergy of the brain and the body that interact with 
the environment \cite{pfeifer2006body}. At the implementation level, \emph{embodied cognition} might be obtained by applying the
several methods that exist
for the multidisciplinary design optimization of hardware and control design \cite{li2001design,allison2013multidisciplinary}.  The key idea here is to consider 
not only the kinematics 
of a mechanism but also its dynamics. Then, 
co-optimization of control and hardware design parameters can be attempted 
via classical optimization techniques, such as block coordinate descendent methods \cite{bertsekas1997nonlinear}. Yet, the specific use-case of human-robot interaction was not investigated using these methods.

Machine learning techniques provide tools that define another route for obtaining embodied cognition. Several works use deep reinforcement learning techniques \cite{ha2019reinforcement}, \cite{chen2020hardware} and evolutionary algorithms \cite{shiakolas2002optimum}, \cite{auerbach2010dynamic}, \cite{bhatia2021evolution} to port the principles of embodied cognition onto the robotics field, showing evidence of how the environment and the resulting tasks directly affect the complexity of the agent. Despite these studies, the principles of embodied cognition were applied mainly to simple robotic platforms composed of a limited number of links and did not consider the human-robot interaction use case.
\begin{figure}[!t]
    \centering	\includegraphics[trim=5.0cm 23.0cm 5.0cm 2.0cm, clip=true, width=0.9\columnwidth]{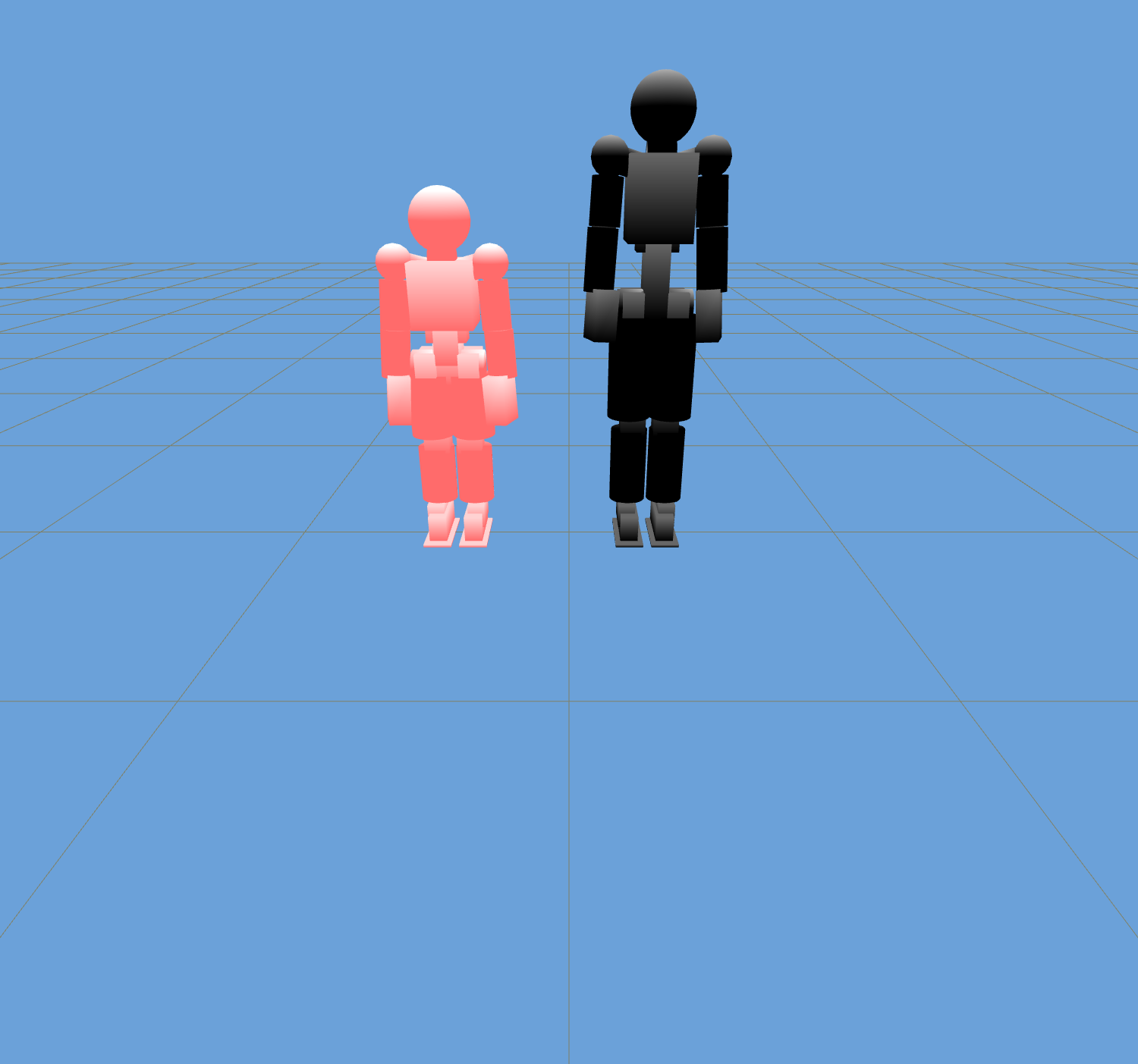}
    \caption{The iCub model (depicted in red), compared to the robot design output of the presented optimization framework.}
    \label{fig:all_results}
\end{figure}

This paper takes the first step towards optimal humanoid robot designs for collaborative tasks while considering the human model in the ergonomics optimization. So, we apply the principles of embodied cognition to humanoid robots tackling human-robot collaborative lifting tasks. This goal is achieved by formulating a physical-consistent parametrization of the humanoid robot kinematics and dynamics with respect to links geometry and density. By using such a parametrization, a non-linear optimization problem is defined considering both human and robot metrics to improve the ergonomics of the interaction.  The proposed methodology has been applied using the iCub humanoid robot \cite{Natale2017} as warm start, and it is the first step towards the design of the ergoCub robot, a humanoid robot developed for ergonomics collaboration with humans in industrial and healthcare environments. The output of the optimization results in the robot design that is depicted in Fig. \ref{fig:all_results}. The optimum design proposed decreases the robot's energy expenditure and preserves the human one. In addition, the proposed robot is able to collaborate with humans in heights in the range of $0.8-1.5 \text{ } \meter $ unlike iCub which is limited to the range of $0.8-1.2\text{ } \meter $.

The remainder of the paper is structured as follows: in Section \ref{sec:background}, notation and modeling used throughout the paper are presented. In Section \ref{sec:hardware-parametrization}, the proposed hardware parametrization is presented along with the associated optimization problem for ergonomics optimization. In Section \ref{sec:validation}, the proposed approach is validated and the associated results are presented and discussed. Conlusions and perspectives close the paper in Section \ref{sec:conclusion}.

\section{BACKGROUND}
\label{sec:background}
\subsection{Notation}
\label{sec:background:notation}
The notation used in this paper is the following: 
\begin{itemize}
    \item $\mathcal{I}$ indicates the inertial reference frame.
    \item $e_i \in \mathbb{R}^m$ is the canonical vector, it is formed by all zeros but the $i^{th}$ component which is equal to 1.
    \item $g$ is the gravitational acceleration norm.
    \item $\mathbb{1}_{n} \in \mathbb{R}^{n \times n}$ denotes an identity matrix.
    \item  $\prescript{\mathcal{A}}{}{p}_{\mathcal{B}} \in \mathbb{R}^3$ is the position of the origin of the frame $\mathcal{B}$ with respect to the frame $\mathcal{A}$.
    \item  $\prescript{\mathcal{A}}{}{R}_{\mathcal{B}} \in SO(3)$ is the rotation matrix of a frame $\mathcal{B}$ with respect to a frame $\mathcal{A}$.
    \item $\prescript{\mathcal{A}}{}{\omega}_{\mathcal{B}} \in \mathbb{R}^3$ is the angular velocity expressed in $\mathcal{A}$ of the frame $\mathcal{B}$ with respect to the frame $\mathcal{A}$.
    \item The $.^{\dagger}$ indicates the Moore-Penrose pseudo-inverse.
    \item The operator $S(.) :\mathbb{R}^{3} \to SO(3)$ denotes \textit{skew-symmetric} vector operator: given two vectors $v,u \in \mathbb{R}^{3}$, it is defined as $v \times u = S(v)u$.
    \item $_AX^{B}$ denotes a wrench 6D vector transform, as defined in \cite{traversaro2016multibody}, such that $_AX^{B}=\begin{bmatrix} {}^{A}R_B & 0 \\ S(p_B - p_A) & {}^{A}R_B \end{bmatrix}$
    \item Given $\mathrm{v} = \begin{bmatrix}v^T & \omega^T\end{bmatrix} \in \mathbb{R}^6$, $\mathrm{v}\bar{\times}^*$ is the 6D dual cross product operator \cite{featherstone2014rigid} defined as: 
        $\mathrm{v}\bar{\times}^*=\begin{bmatrix}
            S(\omega) & 0 \\
            S(v) & S(\omega)
        \end{bmatrix}$
    \item $a^g \in \mathbb{R}^{6}$ is the \emph{proper} body acceleration thus the difference between the body acceleration and the gravity acceleration, i.e. 
    $a^g = \dot{v} - \begin{bmatrix} \prescript{\mathcal{A}}{}{R}_{\mathcal{B}}^Tg \\ 0_{3} \end{bmatrix}$
    \item $z_{\mathcal{C}} \in \mathbb{R}^3$ is the versor of the z-axis of the frame $C$.
    \item The operator $\left\lVert . \right\rVert_2$ indicates vector squared norm. Given $v \in \mathbb{R}^{n}$, it is defined as $\left\lVert v \right\rVert_2 = \sqrt{v_1^2+...+v_n^2}$.
\end{itemize}
 \subsection{Rigid Body Dynamics}
 \label{sec:background-rigid-body-dynamics}
 The  dynamics of a rigid body is defined by the Newton-Euler equations \cite{featherstone2014rigid}, given by: 
 \begin{equation}
 \label{eq:rigidBodyDynamics}
     Ma^g +\mathrm{v}\bar{\times}^*M\mathrm{v} = f, 
 \end{equation}
 where $\mathrm{v} \in \mathbb{R}^6$ is the rigid body velocity, $f \in \mathbb{R}^6$ is the resulting wrench applied to the body, $M \in \mathbb{R}^{6\times6}$ is the 6D inertia matrix defined as: 
 \begin{equation}
 \label{eq:rigidBodyMass}
     M = \begin{bmatrix}m\mathbb{1}_{3\times3} & -mS(c) \\mS(c) & I_\mathcal{A}\end{bmatrix},
 \end{equation}
 with $m \in \mathbb{R}^{+}$ the rigid body mass, $c \in \mathbb{R}^3$ is the center of mass of the rigid body, expressed in the frame $\mathcal{A}$, $I_\mathcal{A} \in \mathbb{R}^{3 \times 3}$ is the  3D rigid body inertia matrix expressed in the frame $\mathcal{A}$.
 
 \subsection{Robot Modelling}
 \label{sec:background-robot-modelling}
 A humanoid robot is a multi-body mechanical system which is composed of $n +1$ rigid bodies, i.e. the \textit{links}, which are connected by $n$ \textit{joints}  with one {degree of freedom} (DoF) each.  The system is considered as {floating base}, therefore none of the links has an {a priori} constant \emph{pose}, i.e. \emph{ position-and-orientation}, with respect to the inertial frame $\mathcal{I}$. The \textit{base frame} is defined as a frame attached to a specific link of the chain, and it is denoted by $\mathcal{B}$.  
The \textit{model configuration} is defined by the pose of the \textit{base frame} together with the \textit{joint positions} and it belongs to the Lie group $\mathbb{Q}=\mathbb{R}^{3+n}\times SO(3) $. An element of the configuration space $q \in \mathbb{Q}$ is defined as the triplet $q = (\prescript{\mathcal{I}}{}{p}_{\mathcal{B}}, \prescript{\mathcal{I}}{}{R}_{\mathcal{B}}, s)$ where $\prescript{\mathcal{I}}{}{p}_{\mathcal{B}} \in \mathbb{R}^3$ and $\prescript{\mathcal{I}}{}{R}_{\mathcal{B}} \in SO(3)$ denote the position and the orientation of the \textit{base frame} respectively, and $s \in \mathbb{R}^n$ is the joints configuration representing the topology of the mechanical system. Given a frame $\mathcal{A}$ attached to the mechanical system, it is possible to compute its position and orientation from the \textit{model configuration} via geometrical forward kinematic map $h_{\mathcal{A}}(\cdot):\mathbb{Q} \to SO(3) \times \mathbb{R}^3$.
The \textit{model velocity}, referred to as $\mathbb{V} = \mathbb{R}^{6+n}$, is composed of the linear and angular velocity of the \textit{base frame} together with the \textit{joint velocities}. An element of the configuration velocity space $\nu \in \mathbb{V}$ is defined as $\nu = (\prescript{\mathcal{I}}{}{\mathrm{v}}_{\mathcal{B}}, \dot{s})$ where $\prescript{\mathcal{I}}{}{\mathrm{v}}_{\mathcal{B}}=(\prescript{\mathcal{I}}{}{\dot{p}}_{\mathcal{B}}, \prescript{\mathcal{I}}{}{\omega}_{\mathcal{B}}) \in \mathbb{R}^6$ denotes the linear and angular velocity of the \textit{base frame}, and $\dot{s}$ denotes the joint velocities. Given frame $\mathcal{A}$ attached to the mechanical system,  it is possible to compute its linear and angular velocity, namely $\prescript{\mathcal{I}}{}{\mathrm{v}}_{\mathcal{A}}=(\prescript{\mathcal{I}}{}{\dot{p}}_{\mathcal{A}}, \prescript{\mathcal{I}}{}{\omega}_{\mathcal{A}})$  via the \textit{Jacobian} ${J}_{\mathcal{A}}={J}_{\mathcal{A}}(\cdot) \in \mathbb{R}^{6 \times (n+6)}$ which maps the system velocity $\nu$ into the frame velocity $\prescript{\mathcal{I}}{}{\mathrm{v}}_{\mathcal{A}}$ i.e.  $\prescript{\mathcal{I}}{}{\mathrm{v}}_{\mathcal{A}} = {J}_{\mathcal{A}}(\cdot)  \nu$.
 
\section{HARDWARE PARAMETRIZATION AND OPTIMUM HARDWARE DESIGN}
\label{sec:hardware-parametrization}
\subsection{Rigid Body Parametrization}
\label{sec:optimumHardware:hardware-parametrization}
The aim of this Section is to identify the relationships between hardware parameters and rigid body dynamics. The rigid body inertial characteristics are described by the body density $\rho \in \mathbb{R}^+$, and the geometry represented with a set of parameters $l \in \mathbb{R}^{n_l}$. The geometry parameters $l$ identify the volume $\mathcal{V} = \mathcal{V}(l) \in \mathbb{R}^3$ which is the region where $\rho \neq 0$. The rigid body mass $m$, the inertia $ I_\mathcal{A}$ computed with respect to the rigid body frame $\mathcal{A}$, and the rigid body center of mass $c$ are function of $\rho$ and $l$ via the equations: 

\begin{IEEEeqnarray}{RLC}
\IEEEyesnumber
&m(l, \rho) = \int \int \int_{\mathcal{V}(l)}\rho(r)\cdot dr, \IEEEyessubnumber\label{eq:shapeMass}\\
&I_{\mathcal{A}}(l, \rho) = - \int \int \int_{\mathcal{V}(l)} \rho(r)\left[ S(r)\right ]^2\cdot dr, \IEEEyessubnumber\label{eq:shapeInertia} \\ 
&c (l, \rho) = \frac{\int \int \int_{\mathcal{V}(l)}r\rho(r) \cdot dr}{m(l,\rho)}. \IEEEyessubnumber \label{eq:shapeCoM}
\yesnumber
\end{IEEEeqnarray}

Since the quantities in Eq. \eqref{eq:shapeMass}, \eqref{eq:shapeInertia}, \eqref{eq:shapeCoM} are obtained from the link density and geometric properties, the physical consistency is ensured, and we can rewrite Eq. \eqref{eq:rigidBodyDynamics} as a function of the  hardware parameters: 
 \begin{equation}
 \label{eq:rigidBodyDynamicsWithHardw}
     M(l, \rho)a^g +\mathrm{v}\bar{\times}^*M(l, \rho)\mathrm{v} = f. 
 \end{equation}
 
\subsection{Multi Rigid Body Parametrization}
\label{sec:hardware-optimization-multi-rigid-body}
The same parametrization of Eq. \eqref{eq:rigidBodyDynamicsWithHardw} can be performed for each link of the robot kinematic chain, and it is possible to define ${\pi}$ as the set of hardware parameters associated with the robot links. The robot dynamics is then formulated by applying the Euler-Poincarè formalism \cite{Marsden2010} to the kinematic chain parametrized with respect to the hardware parameters and is described by a set of differential equations together with holonomic constraints describing the interaction with the environment as 
\begin{IEEEeqnarray}{RLC}
\IEEEyesnumber
\label{eq:constrained-dynamic-hardware-param}
& M(q, \pi) \dot{\nu} + h(q,\nu,\pi) = B {\tau} + J_c^T(q,\pi) f, \IEEEyessubnumber \label{eq:constrained-dynamic-hardware-param:dynamics} \\
& \dot{J_c}(q,\pi) \nu + J_c(q,\pi) \dot{\nu} = 0,\IEEEyessubnumber  \label{eq:constrained-dynamic-hardware-param:constraints} 
\yesnumber
\end{IEEEeqnarray}
with $M \in \mathbb{R}^{n+6 \times n+6}$ the mass matrix, the term $h \in \mathbb{R}^{n+6}$ accounting for Coriolis and gravity forces, $B = (0_{n \times 6},\mathbb{1}_n)^T$ is a selector matrix, ${\tau}  \in \mathbb{R}^{n}$ is a vector representing the robot's joint torques, $f \in \mathbb{R}^{6n_c}$ represents the wrenches acting on $n_c$ contact links of the robot, and $J_c \in \mathbb{R}^{n+6 \times 6n_c}$ is the Jacobian of the contact frames. It is worth noticing as also the Jacobian is a function of the hardware parameters, even though it is a purely kinematic function. Indeed, changes in the length of the links, thus in $\pi$, will result in a different system kinematic, hence a different Jacobian relationship. 
The robot dynamics of Eq. \eqref{eq:constrained-dynamic-hardware-param:dynamics}, can be projected into the holonomic constraint \eqref{eq:constrained-dynamic-hardware-param:constraints}, obtaining the following relationship:  
\begin{align}
\label{eq:constrained-dynamic-projected}
\begin{split}
& N_{\Lambda}(q, \pi) [M(q, \pi)\dot{\nu} + h(q, \pi,\nu)-B\tau]=0,
\end{split}
\end{align}
with $N_{\Lambda}(q,\pi)$ defined as 
\begin{equation}
\label{eq:N_A}
N_{\Lambda}=\mathbb{1}-J_c^T\left(J_c M^{-1}J_c^T\right)^{-1}J_c M^{-1},
\end{equation}
where, for the sake of clarity, the dependency on $q$ and $\pi$ has been omitted. 
\subsection{Human-Robot Coupled Dynamics Parametrization}
\label{sec:hardwareParametrization-Human-robot-dynamics}
In the scenario addressed by this paper, we consider human-robot collaborative lifting tasks, where a human being and a robot interact by holding together a payload. In this scenario, the contacts are not limited to the ones with the environment, namely the ground, but also the interaction with the payload should be taken into account. 
The coupled human-robot-payload dynamics can be described as exposed in \cite{tirupachuri2019}, where the human is modeled as a multi-rigid body floating base system. Starting from Section \ref{sec:hardware-optimization-multi-rigid-body}, the dependency on the robot hardware parameters has been made explicit, leading to the following
\begin{equation}
\begin{split}
\label{eq:multi-system-equations}
& \begin{bmatrix} M_1(q_1, \pi) & 0 & 0\\\ 0 & M_2(q_2) & 0 \\ 0 & 0 & M_3(q_3)\end{bmatrix} \begin{bmatrix} \dot{\nu}_1 \\ \dot{\nu}_2 \\ \dot{\nu}_3 \end{bmatrix} + \begin{bmatrix} h_1(q_1,\nu_1,\pi) \\ h_2(q_2,\nu_2) \\h_3(q_3, \nu_3) \end{bmatrix} =\\ & \begin{bmatrix} B_1 & 0 \\\ 0 & B_2 \\\ 0& 0 \end{bmatrix}  \begin{bmatrix} \tau_1 \\\ \tau_2 \end{bmatrix} + \mathbf{Q}(q_1,q_2,q_3, \pi)^T \mathbf{f}, \\
& {\mathbf{\dot{Q}}}(q_1,q_2,q_3,\pi) \begin{bmatrix} \nu_1 \\ \nu_2 \\ \nu_3 \end{bmatrix} +\mathbf{Q}(q_1,q_2,q_3,\pi) \begin{bmatrix} \dot{\nu}_1 \\ \dot{\nu}_2 \\ \dot{\nu}_3 \end{bmatrix} = 0,
\end{split}
\end{equation}
where the terms related to the human and the robot are referred to with, respectively,  the subscript $1$, $2$, and the payload quantities are referred to with the subscript $3$. The composite matrices are identified with  $\mathbf{bold}$. In Eq. \eqref{eq:multi-system-equations},  $\mathbf{Q}$ is a coupling matrix considering both the constraints of the contacts with the environment ($J^{e}\nu=0$) and those of the agent-payload contact points for each agent, namely the human and the robot, ($J_1^{i}\nu_1=J_3^{i}\nu_3$, $J_2^{i}\nu_1=J_3^{i}\nu_3$), and $\mathbf{f}$ is a vector containing all the interaction wrenches (exchanged with the environment and between the agents and the payload) taking into account the action-reaction property for internal forces ($f^{i}_1=-f^{i}_3$ $f^{i}_2=-f^{i}_3$) and reflecting the ordering in the constraints matrix $\mathbf{Q}$. Then Eq. \eqref{eq:multi-system-equations} can be written in its compact form:
\begin{equation}
\label{eq:multi-system-equations-compact}
\begin{split}
& \mathbf{M} (\boldsymbol{q}, \pi){\boldsymbol{\dot{\nu}}} + \mathbf{h}(\boldsymbol{q}, \boldsymbol{\nu},\pi) = \mathbf{B} \boldsymbol{\tau} + \mathbf{Q}^T(\boldsymbol{q}, \pi) \mathbf{f}, \\
& {\mathbf{\dot{Q}}}(\boldsymbol{q}, \pi) \boldsymbol{\nu} +\mathbf{Q}(\boldsymbol{q}, \pi) \boldsymbol{\dot{\nu}} = 0.
\end{split}
\end{equation}
Also in this case the projection of the dynamics into the constraints can be formulated as 
\begin{align}
\label{eq:constrained-dynamic-projected-multi-agent}
\begin{split}
& \boldsymbol{N}_{\Lambda}(\boldsymbol{q}, \pi)[\boldsymbol{M}(\boldsymbol{q}, \pi){\boldsymbol{\dot{\nu}}} + \boldsymbol{h}(\boldsymbol{q}, \boldsymbol{\nu},\pi)-\boldsymbol{B}\boldsymbol{\tau}
]=0,
\end{split}
\end{align}
Where $\boldsymbol{N}_{\Lambda}$ is defined as 
\begin{equation}
\boldsymbol{N}_{\Lambda}(\boldsymbol{q}, \pi)=\mathbb{1}-\boldsymbol{Q}^T\left(\boldsymbol{Q} \boldsymbol{M}^{-1}\boldsymbol{Q}^T\right)^{-1}\boldsymbol{Q} \boldsymbol{M}^{-1},
\end{equation}
 Where, for the sake of clarity, the dependency on $\boldsymbol{q}$ and $\pi$ has been omitted, hence $\boldsymbol{Q} = \boldsymbol{Q}(\boldsymbol{q}, \pi)$ and $\boldsymbol{M} =\boldsymbol{M}(\boldsymbol{q}, \pi) $
\begin{figure*}[!t]
\centering
\begin{subfigure}[b]{0.23\textwidth}
 \includegraphics[trim=14cm 20cm 14cm 0cm, clip=true,width=0.98\columnwidth]{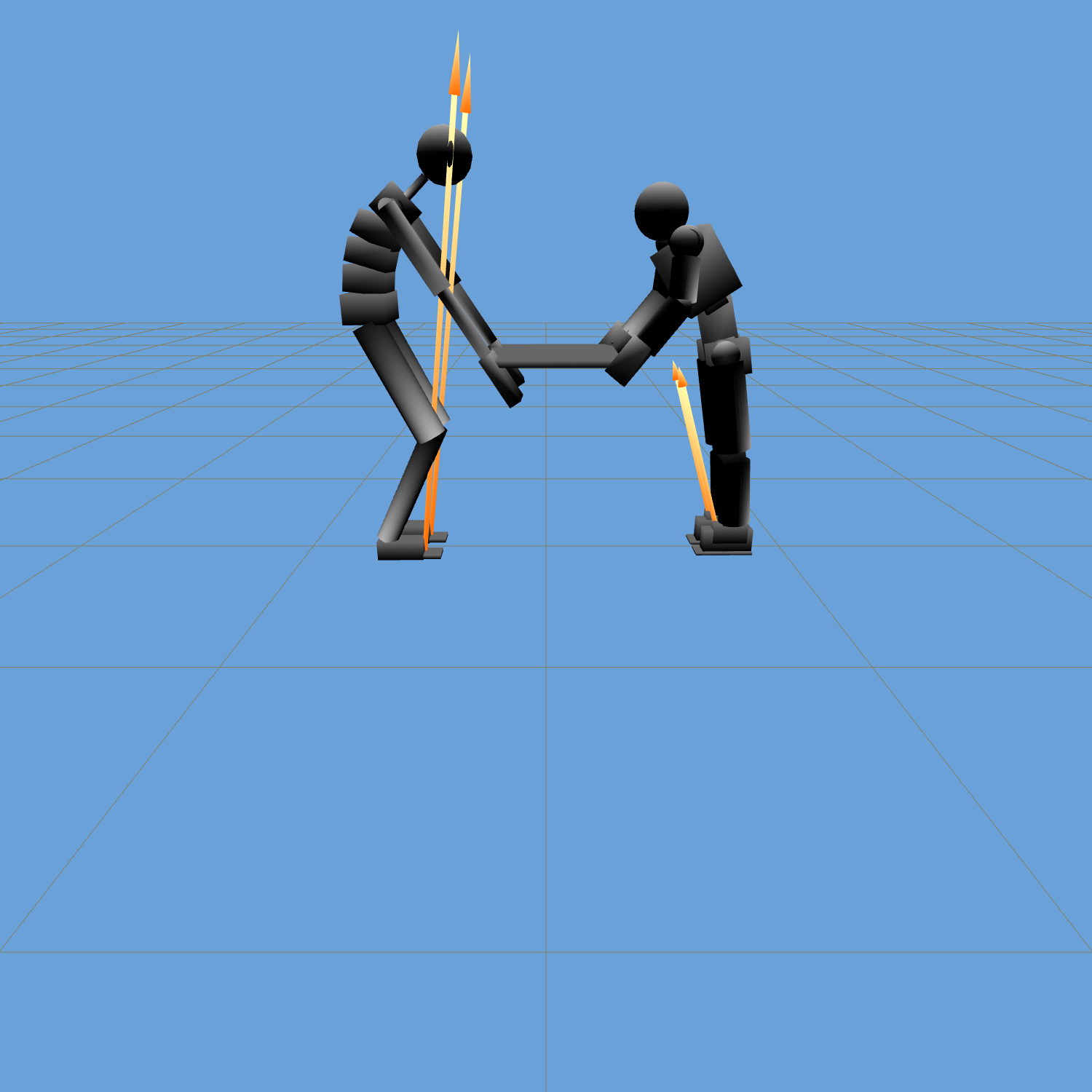}
 \caption{}
 \label{fig:optimized_0}
\end{subfigure}
\begin{subfigure}[b]{0.23\textwidth}
 \includegraphics[trim=14cm 20cm 14cm 0cm, clip=true,width=0.98\columnwidth]{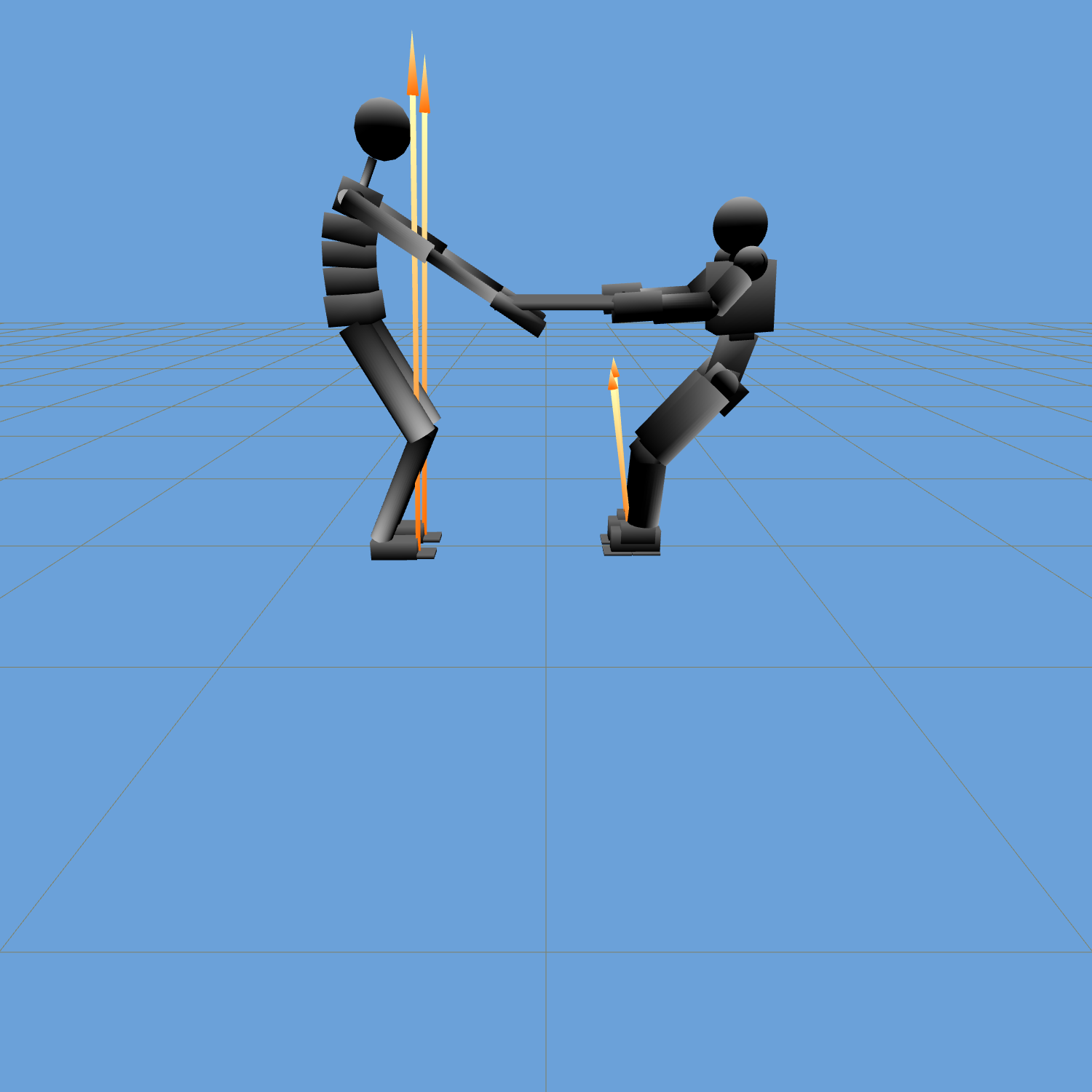}
 \caption{}
 \label{fig:optimized_1}
\end{subfigure}
\begin{subfigure}[b]{0.23\textwidth}
 \includegraphics[trim=14cm 20cm 14.0cm 0cm, clip=true,width=0.98\columnwidth]{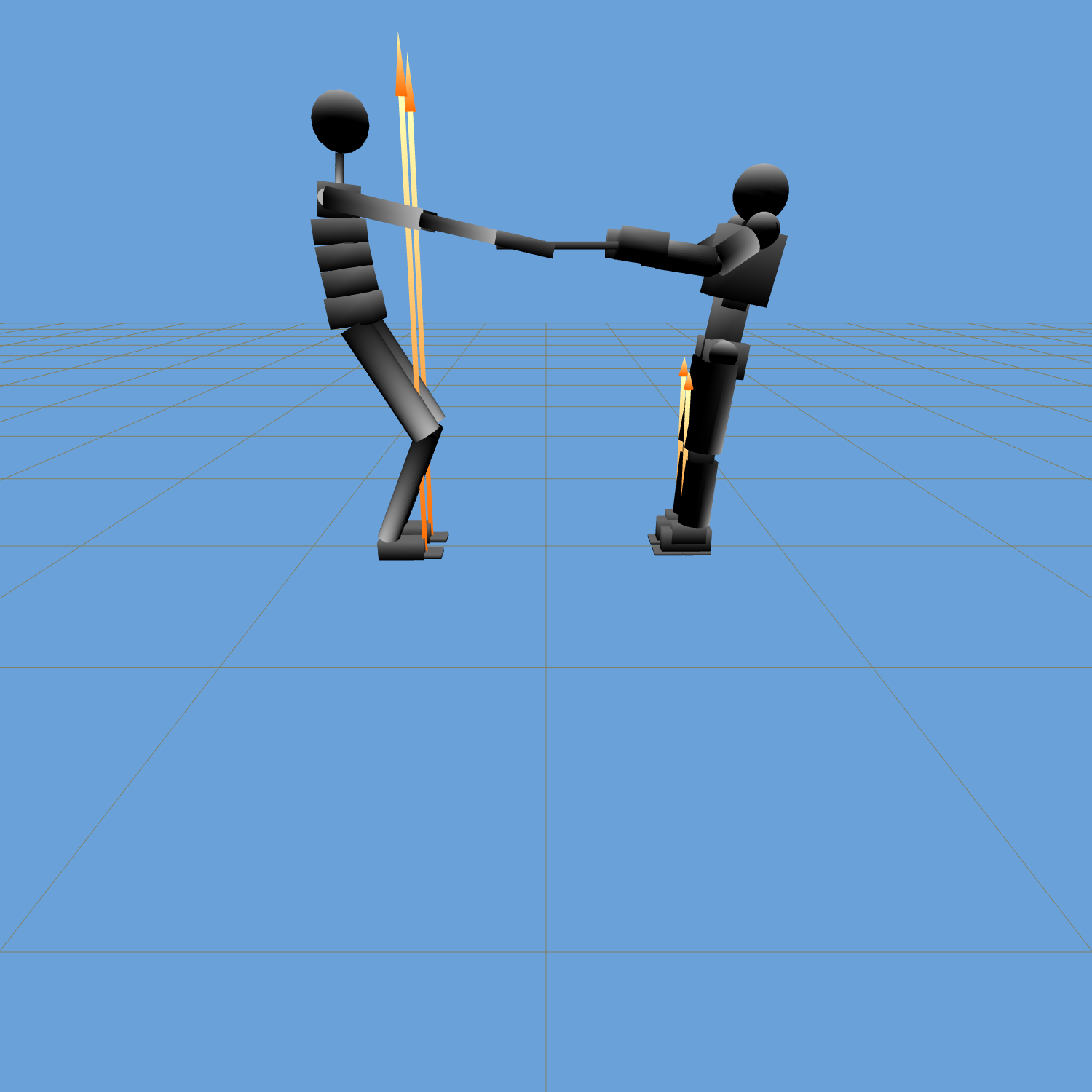}
 \caption{}
 \label{fig:optimized_2}
\end{subfigure}
\begin{subfigure}[b]{0.23\textwidth}
 \includegraphics[trim=14cm 20cm 14cm 0cm, clip=true,width=0.98\columnwidth]{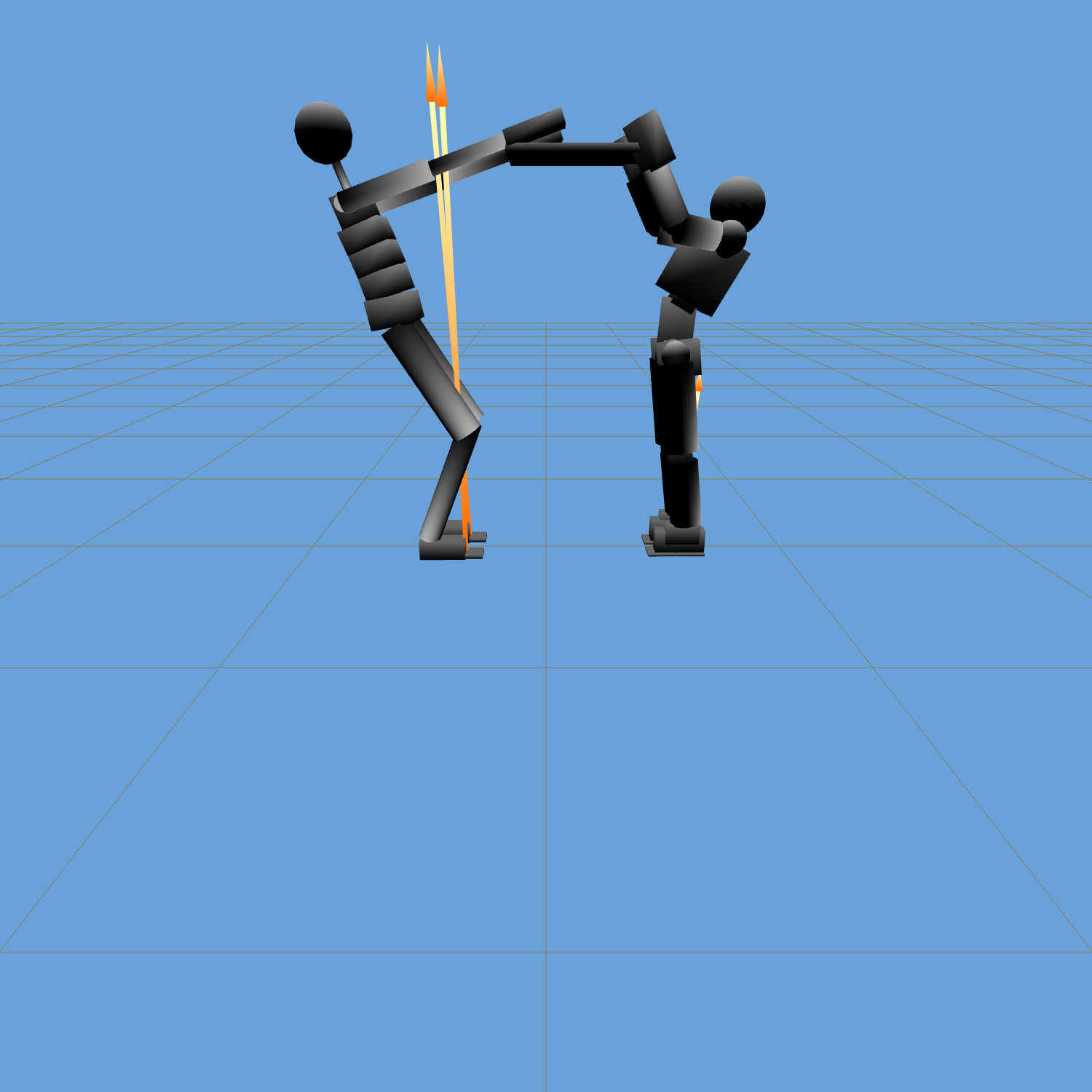}
 \caption{}
 \label{fig:optimized_3}
\end{subfigure}

\caption{Optimized robot collaborating with the human. In (a) the load is at an height of $0.8 \textit{ } \meter$. In (b) the load is at an height of $1.0 \textit{ } \meter$. In (c) the load is at an height of $1.2 \textit{ } \meter$. In (d) the load is at an height of $1.5 \textit{ } \meter$. In orange, the wrenches exchanged with the ground are visualized.}
\label{fig:optimized}
\end{figure*}

\begin{figure*}[!t]
\centering
\begin{subfigure}[b]{0.23\textwidth}
 \includegraphics[trim=14cm 20cm 14cm 0cm, clip=true,width=0.98\columnwidth]{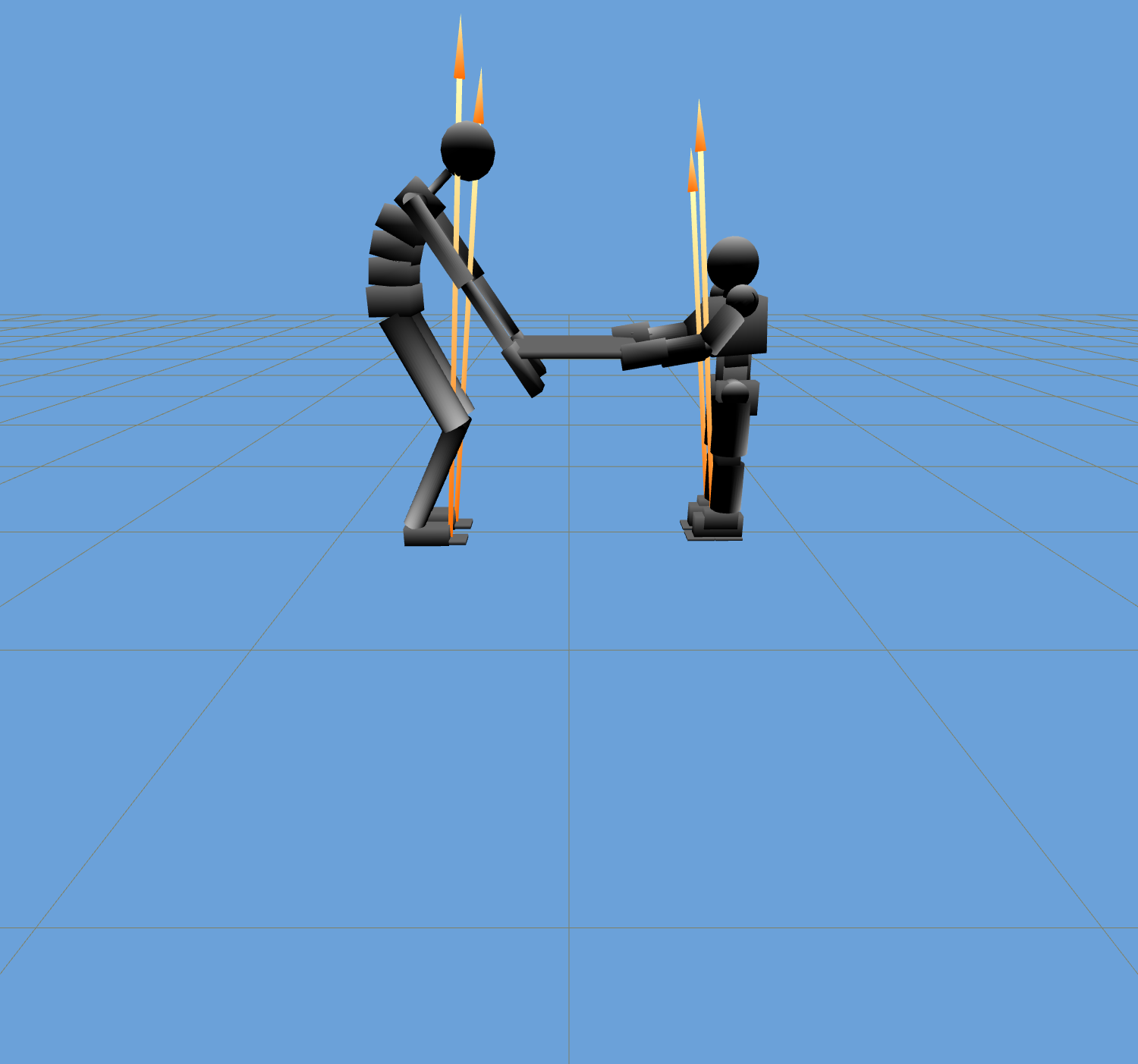}
 \caption{}
 \label{fig:original_0}
\end{subfigure}
\begin{subfigure}[b]{0.23\textwidth}
 \includegraphics[trim=14cm 20cm 14cm 0cm, clip=true,width=0.98\columnwidth]{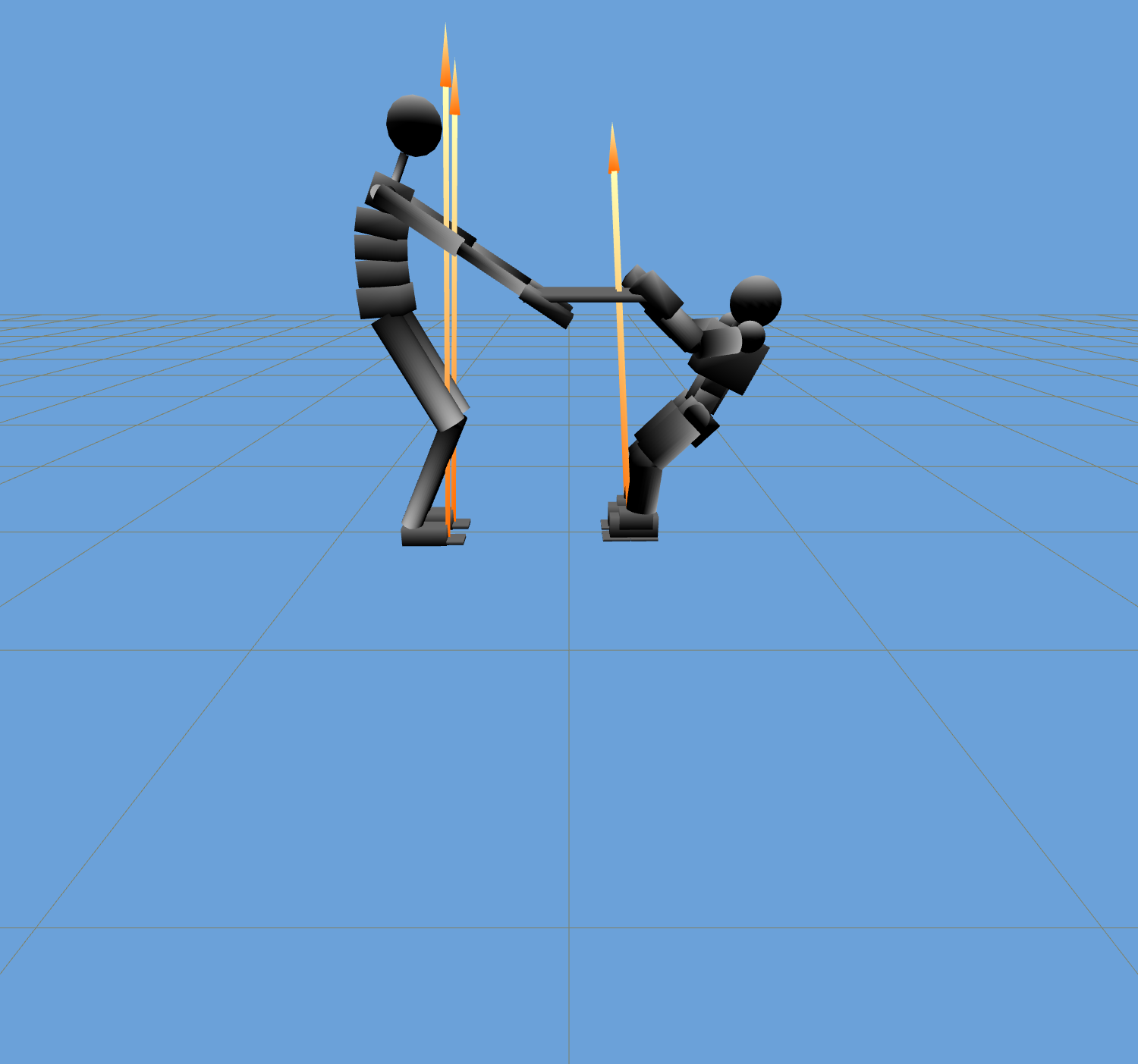}
 \caption{}
 \label{fig:original_1}
\end{subfigure}
\begin{subfigure}[b]{0.23\textwidth}
 \includegraphics[trim=14cm 20cm 14cm 0cm, clip=true,width=0.98\columnwidth]{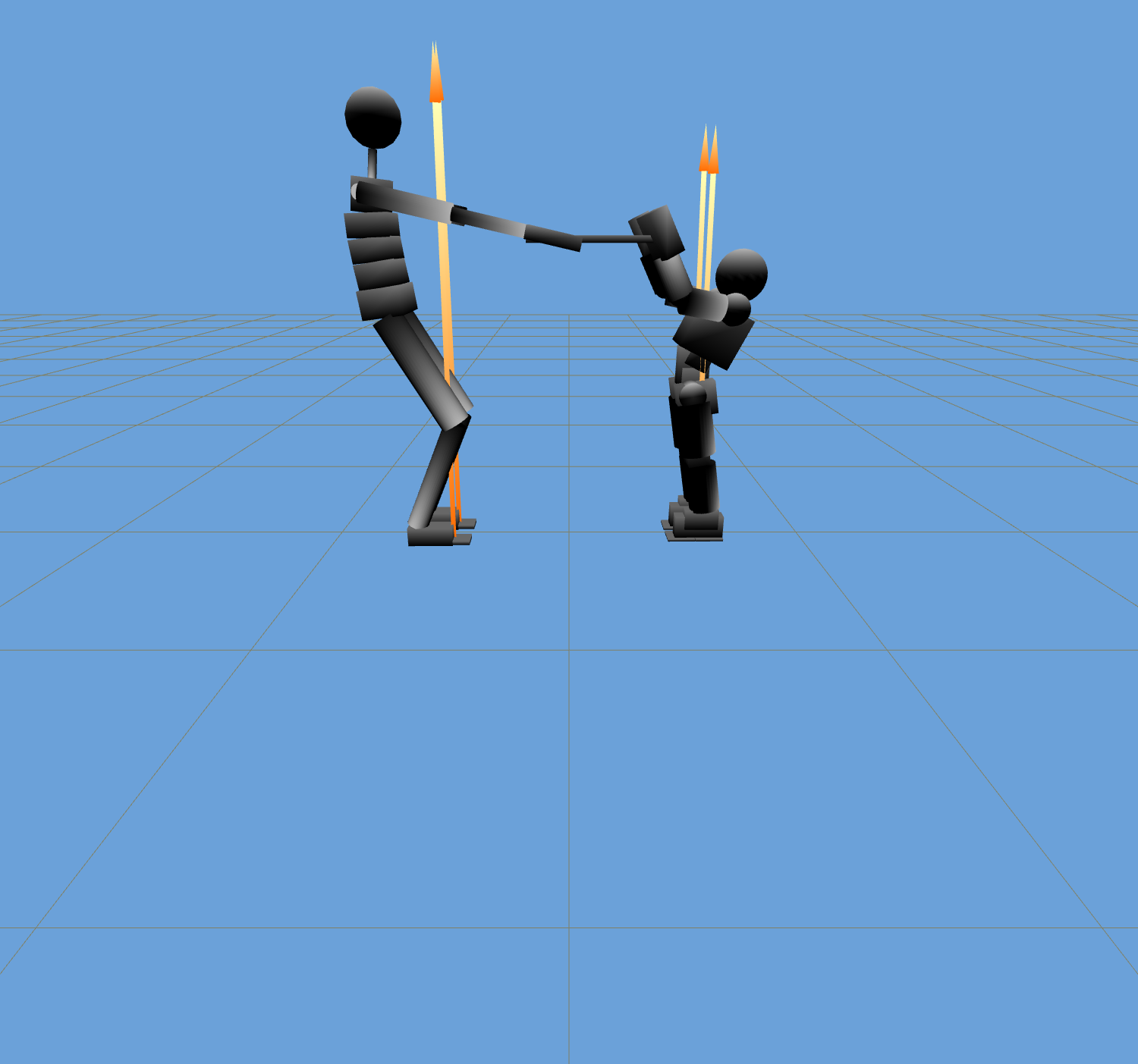}
 \caption{}
 \label{fig:original_2}
\end{subfigure}
\caption{The original robot collaborating with the human. In (a) the load is at an height of $0.8 \textit{ } \meter$. In (b) the load is at an height of $1.0 \textit{ } \meter$. In (c) the load is at an height of $1.2 \textit{ } \meter$. In orange, the wrenches exchanged with the ground are visualized.}
\label{fig:original}
\end{figure*}
\subsection{Ergonomics}
\label{sec:optimumHardware:ergonomicDesign}
As highlighted in Section \ref{sec:introduction}, when a human interacts and collaborates with a robot, an optimal interaction from the ergonomic point of view should be targeted \cite{Rapetti2021}. The term \emph{ergonomics} refers to the discipline of designing methods aimed at optimizing human well-being and performance during the interaction with the environment \cite{dul2012strategy}. In this respect, one metric considered is the expenditure of energy represented by the robot's and human's internal torques. Such analysis exploits the joint torques redundancy associated with a static equilibrium configuration \cite{matheson1959hyperstatic} to ensure a safe and efficient interaction: the \emph{force ergonomics} aims at minimizing the joint torques at equilibrium configuration through contact forces optimization while the \emph{postural ergonomics} exploits the system configuration to optimize the joint torques and achieve an ergonomic interaction \cite{Rapetti2021}. Thanks to the parametrization of Section \ref{sec:hardware-parametrization}, it is possible to consider the influence of the hardware parameters $\pi$ on the ergonomics of the interaction. 
In this regard, the joint torques can be evaluated as a metric of the \emph{postural optimization}, analogously to the work done in \cite{rapetti2021shared}.
Starting from the coupled dynamic projected in the null space of Eq. \eqref{eq:constrained-dynamic-projected-multi-agent}, it is possible to compute the joint torques (both of the human and the robot) at static configuration, as a function of the agent's configurations and the robot parameters as per the following Eq.
\begin{equation}
\label{eq:tau_on_parameters}
 \boldsymbol{\tau}(\mathbf{q}, \pi) =\left(\boldsymbol{N}_{\Lambda}(\mathbf{q}, \pi)\boldsymbol{B}\right)^{\dagger}\boldsymbol{N}_{\Lambda}(\mathbf{q}, \pi)\boldsymbol{g}(\mathbf{q},\pi).    
\end{equation}
The interaction wrenches depends on the hardware parameters, indeed, starting from Eq. \eqref{eq:constrained-dynamic-projected-multi-agent}, the relationship of Eq. \eqref{eq:wrenches_on_parameters} can be obtained. Where $\tau$ is computed as per Eq. \eqref{eq:tau_on_parameters}.
 \begin{equation}
 \label{eq:wrenches_on_parameters}
 \boldsymbol{f}(\mathbf{\mathbf{q}, \pi}) = \left[\boldsymbol{Q}\boldsymbol{M}^{-1}\boldsymbol{Q}^T\right]^{-1} \left[\boldsymbol{Q} \boldsymbol{M}^{-1}\left(-\boldsymbol{B}\boldsymbol{\tau}  +\boldsymbol{g}\right) \right]. \end{equation}
 Note that in Eq. \eqref{eq:wrenches_on_parameters} the dependency on $\mathbf{q}$ and $\pi$ has been omitted for the sake of clarity. 
\subsection{Ergonomic Design: Optimization Problem}
\label{sec:optimumHardware:optimizationProblem}
\begin{table*}[!t]
\begin{center}
\caption{Parametrization of link inertial characteristics for basic geometrical shapes.}
\label{tab:linkParam}
\begin{tabular}{|c|c|c|c|}
\hline
& \textbf{Sphere} & \textbf{Cylinder} & \textbf{Box}\\ \hline
$m(\pi)$ & $\frac{4\pi (r l_m)^3}{3}\rho$  &$ \pi r^2 (h  l_m)\rho$ & $w h(d l_m) \rho$ \\ \hline 
$I_c(\pi)$ & $m\begin{bmatrix}\frac{2}{5} (r  l_m)^2 & 0 & 0  \\
         0 &\frac{2}{5}(r l_m)^2 & 0 \\
         0 & 0 & \frac{2}{5}(r l_m)^2\end{bmatrix}$ &m$\begin{bmatrix}         \frac{(3r^2+(h l_m)^2)}{12} & 0 & 0  \\
         0 & \frac{(3r^2+(h  l_m)^2)}{12} & 0 \\
         0 & 0 & \frac{1}{2}r^2\end{bmatrix}$ & $m\begin{bmatrix}\frac{h^2+(d l_m)^2}{12} & 0 & 0  \\
         0 & \frac{w^2+(d l_m)^2}{12} & 0 \\
         0 & 0 & \frac{(w^2+h^2)}{12}\end{bmatrix}$ \\
\hline
\end{tabular}
\end{center}
\end{table*}
 As stated in the Section \ref{sec:introduction}, the scenario addressed by this work concerns payload lifting task, where a human and a robot have to collaborate to lift a load at a certain given height $h^*$.  To achieve a human-robot ergonomic interaction, we should minimize the joint torques as discussed in Section \ref{sec:optimumHardware:ergonomicDesign}. Therefore, it is possible to define a non-linear optimization problem to identify $\boldsymbol{q}^*$ and $\pi^*$ that are associated with minimum torque. For the optimization problem, the following objectives have been defined: 
 \begin{itemize}
     \item \textbf{Ergonomic interaction}: to achieve an ergonomic interaction, we should minimize the joint torques, i.e. $t_1=\left\rVert \boldsymbol{\tau} (\boldsymbol{q}, \pi)\right\rVert_2^2 $.
     \item \textbf{Desired densities}: in the set of hardware parameters considered, the links length and density are included. A set of preferable densities, associated with different materials, is defined as $\{\rho^*_1,\rho^*_2,\ldots \}$. A task, driving the density of the $i^{th}$ link towards those values is defined as $t_2= \Pi_i|\rho^*_i-\rho_j|$.
     \item \textbf{Center of pressure}: to ensure the dynamical feasibility of the interaction, both agents should have the feet center of pressure in a stable desired position $CoP^*$. The feet center of pressure can be computed, starting from Eq. \eqref{eq:wrenches_on_parameters} as $CoP= [-\frac{\tau_y}{f_z},  \frac{\tau_x}{f_z}]$. Therefore a task is defined to drive the feet center of pressure towards a stable desired value, i.e.  $t_3=\left\rVert CoP_i-CoP_i^*\right\rVert_2^2 \forall i \text{ } \in [1,2] $. 
     \item \textbf{Center of mass height}: a widespread principle in humanoid robot hardware design consists in maximizing the robot center of mass height to decrease the system bandwidth, hence having a robot more robust w.r.t. less reactive control architectures. For this reason, the following task is defined $t_4 = \left\rVert\frac{1}{\prescript{\mathcal{I}}{}{p}_{z0,com}}\right\rVert_2^2$ where $\prescript{\mathcal{I}}{}{p}_{z0,com}$ is the robot center of mass height computed at null configuration. 
 \end{itemize} 
 Since in a real scenario, robots and humans are generally requested to interact with loads positioned at different heights, a set of target heights $ \boldsymbol{h^*} = \{ h^*_1,h^*_2, \ldots, h^*_{n_k} \} \in \mathbb{R}^{n_k}$ is considered, and the optimization problem of Eq. \eqref{eq:optimizationProblem} has been defined. The search variable of such a problem is defined as $\boldsymbol{y} =     \begin{bmatrix}
\tilde{\boldsymbol{q}} &
\pi
\end{bmatrix}^T$
where  $\tilde{\cdot}$ represents a set of variables, one per target object height considered.
\begin{IEEEeqnarray}{RCL}
\IEEEyesnumber
\label{eq:optimizationProblem}
\boldsymbol{y} ^{*} & = & \underset{\boldsymbol{y}}{\text{argmin}}\left(\sum_{n_k}\left( W_1t_1{+}W_3t_3\right){+}W_2t_2{+}W_4t_4\right)  \nonumber\\
& s.t. & \nonumber\\ 
&&  e_{3}^T \cdot z_{3,k} {=}1 \text{ } \forall k \in \boldsymbol{K}, \IEEEyessubnumber \label{opt:LoadOrientation}\\
&& \prescript{\mathcal{I}}{}{p}_{z,3,k} = h_k^* \text{ }  \forall k \in \boldsymbol{K}, \IEEEyessubnumber \label{opt:LoadHeight}\\
&& \prescript{\mathcal{I}}{}{p}_{LH_i,k}{=}\prescript{\mathcal{I}}{}{p}_{LH_i,3,k} \text{ } \forall i \in \boldsymbol{I}, \text{ }  \forall k \in \boldsymbol{K}, \IEEEyessubnumber \label{opt:LeftHand}\\
&& \prescript{\mathcal{I}}{}{p}_{RH_i,k}{=}\prescript{\mathcal{I}}{}{p}_{RH_i,3,k} \text{ } \forall i \in \boldsymbol{I}, \text{ }  \forall k \in \boldsymbol{K}, \IEEEyessubnumber \label{opt:RightHand}\\
&& \prescript{\mathcal{I}}{}{p}_{z,LF_i,k} = 0 \text{ }  \forall i \in \boldsymbol{I}, \text{ } \forall k \in \boldsymbol{K},  \IEEEyessubnumber \label{opt:FeetPosition-Left}\\
&& \prescript{\mathcal{I}}{}{p}_{z,RF_i,k} = 0 \text{ } \forall i \in \boldsymbol{I}, \text{ }  \forall k \in \boldsymbol{K}, \IEEEyessubnumber \label{opt:FeetPosition-Right}\\
&& e_{3}^T \cdot z_{LF_i,k} = 1 \text{ } \forall i \in \boldsymbol{I},\text{ }  \forall k \in \boldsymbol{K}, \IEEEyessubnumber \label{opt:FeetOrientation-Left}\\
&& e_{3}^T \cdot z_{RF_i,k} = 1 \text{ } \forall i \in \boldsymbol{I}, \text{ }  \forall k \in \boldsymbol{K}. \IEEEyessubnumber \label{opt:FeetOrientation-Left}\\
\nonumber
\end{IEEEeqnarray}
Where $W_1, W_2,W_3,W_4 \in \mathbb{R}$ are the task weights, $\boldsymbol{I}= [1,2]$ and $\boldsymbol{K} = [1,n_k]$.  
In the optimization problem of Eq. \eqref{eq:optimizationProblem}, different constraints have been introduced, to ensure the feasibility of the interaction:
 \begin{itemize}
     \item \textbf{Load orientation}: the load should be kept parallel to the ground, resulting in constraint of Eq. \eqref{opt:LoadOrientation}, where $e_{3}^T \cdot z_{3}$ represents the misalignment in between the gravity direction and the z versor of the load frame.
     \item \textbf{Load height}: the load should be kept at the desired height $h^*$. For this reason, the constraint of Eq. \eqref{opt:LoadHeight}  has been defined, where $\prescript{\mathcal{I}}{}{p}_{z,\mathcal{A}}$ represents the \textit{z}-component of the position of the frame $\mathcal{A}$. 
     \item \textbf{Hand position}: the hands of the robot and the human, should be at the correct position on  the load, this results in the constraint of Eq. \eqref{opt:LeftHand}  and Eq. \eqref{opt:RightHand} where $LH_i$ stands for the left hand frame of the $i^{th}$ agent, $RH_i$ stands for the right hand frame of the $i^{th}$ agent and $\prescript{\mathcal{I}}{}{p}_{\mathcal{A}_i,3}$ stands for the box position for the $\mathcal{A}$ frame of the $i^{th}$ agent. 
     \item \textbf{Feet position}: both the human and the robot feet should be on the ground, resulting in the constraints of Eq. \eqref{opt:FeetPosition-Left}  and Eq. \eqref{opt:FeetPosition-Right}. where $LF_i$ stands for the left foot frame of the $i^{th}$ agent, $RF_i$ stands for the right foot frame of the $i^{th}$ agent.
     \item \textbf{Feet orientation}: both the human and the robot feet should be parallel to the ground, resulting in constraints of Eq. \eqref{opt:FeetPosition-Left} and Eq.\eqref{opt:FeetPosition-Right}. 
 \end{itemize}
\begin{figure*}[t!]
\begin{subfigure}[t]{1.0\columnwidth}
  \includegraphics[trim=0.0cm 0.0cm 0.0cm 0.0cm, clip=true,width=1.0\columnwidth]{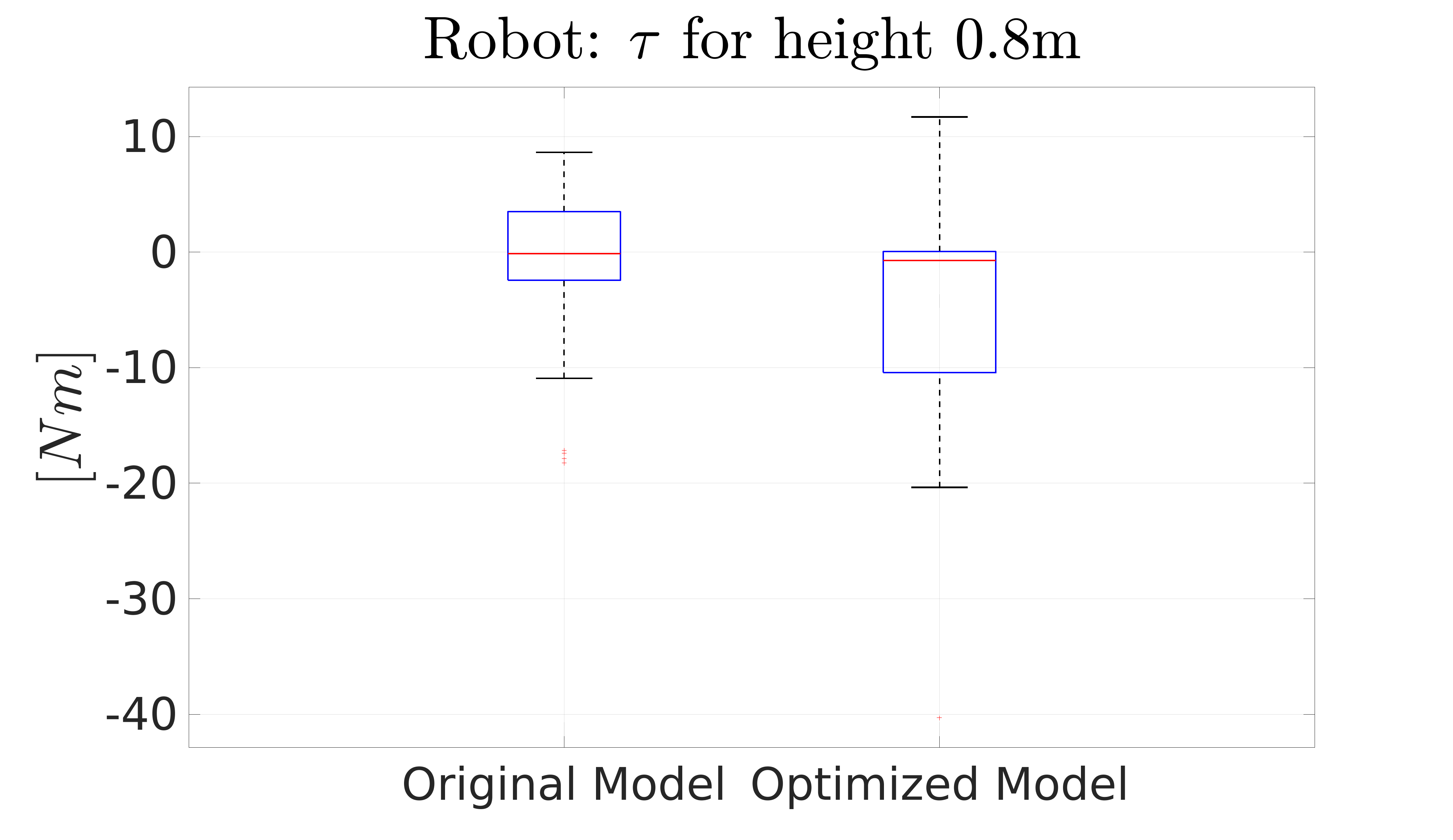}
  \caption{}
  \label{fig:robotTorque_1}
\end{subfigure}
\begin{subfigure}[t]{1.0\columnwidth}
  \includegraphics[trim=0.0cm 0.0cm 0.0cm 0.0cm, clip=true,width=1.0\columnwidth]{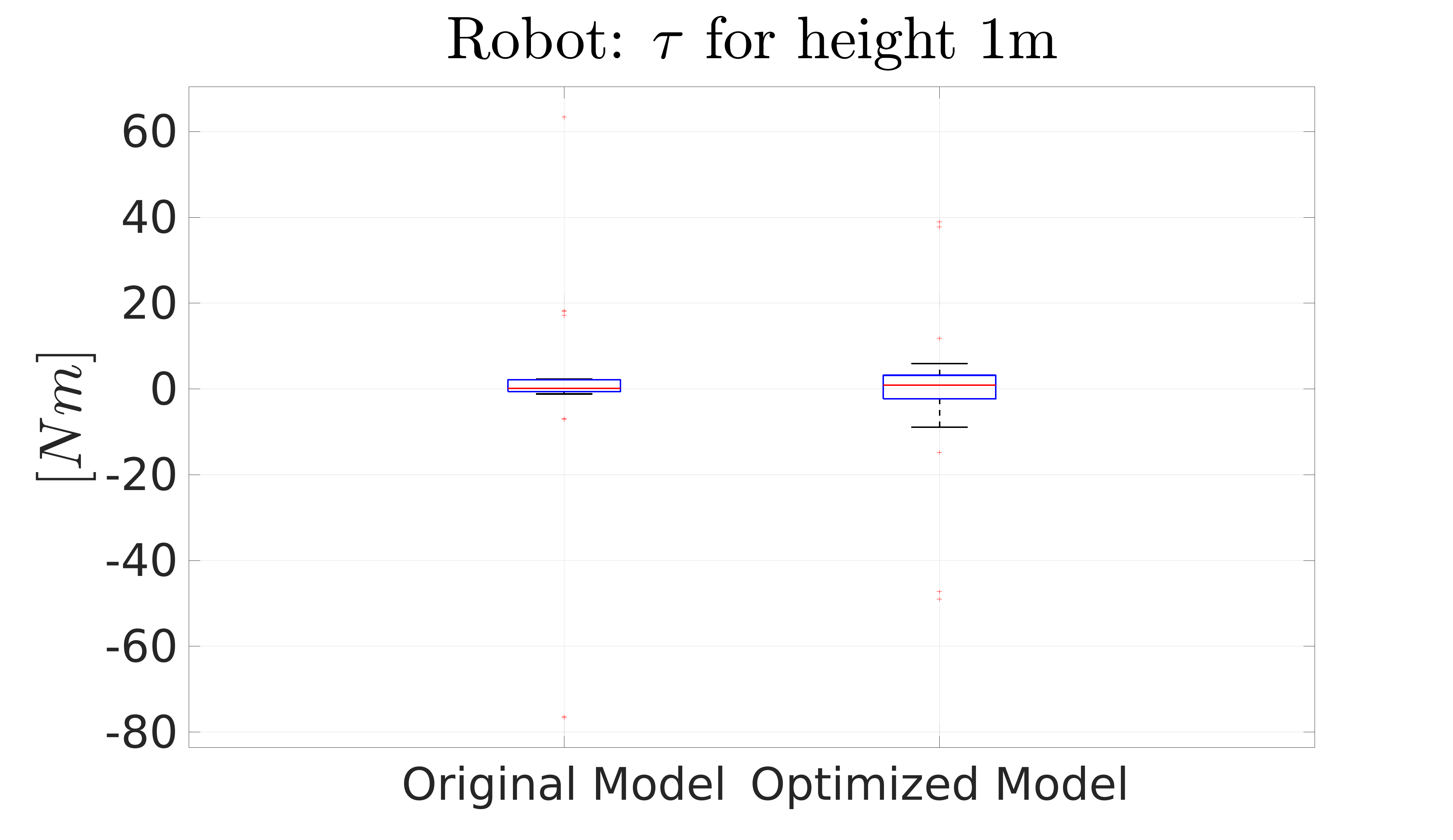}
  \caption{}
  \label{fig:robotTorque_2}
\end{subfigure}
\begin{subfigure}[t]{1.0\columnwidth}
  \includegraphics[trim=0.0cm 0.0cm 0.0cm 0.0cm, clip=true,width=1.0\columnwidth]{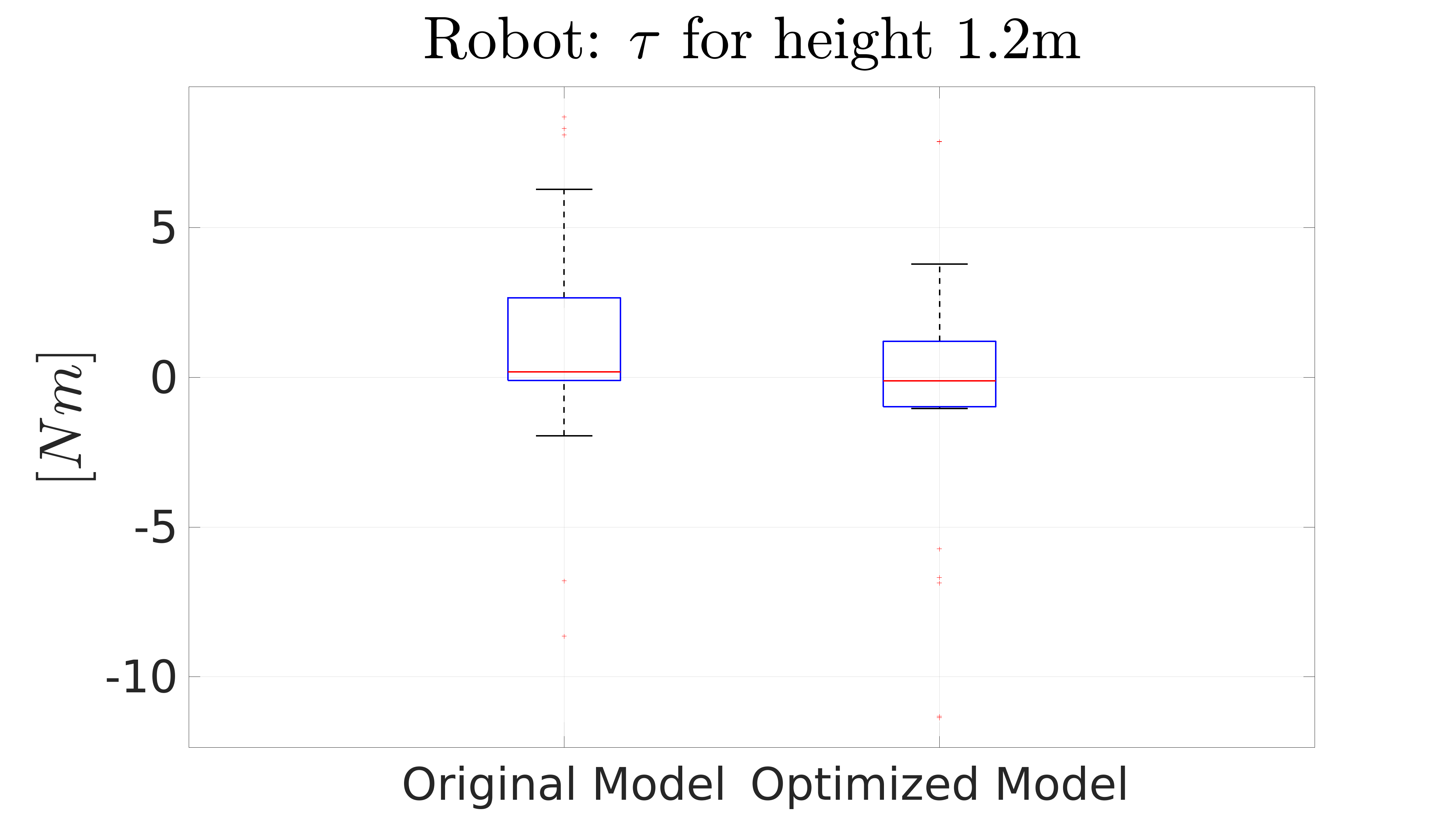}
  \caption{}
  \label{fig:robotTorque_3}
\end{subfigure}
\begin{subfigure}[t]{1.0\columnwidth}
  \includegraphics[trim=0.0cm 0.0cm 0.0cm 0.0cm, clip=true,width=1.0\columnwidth]{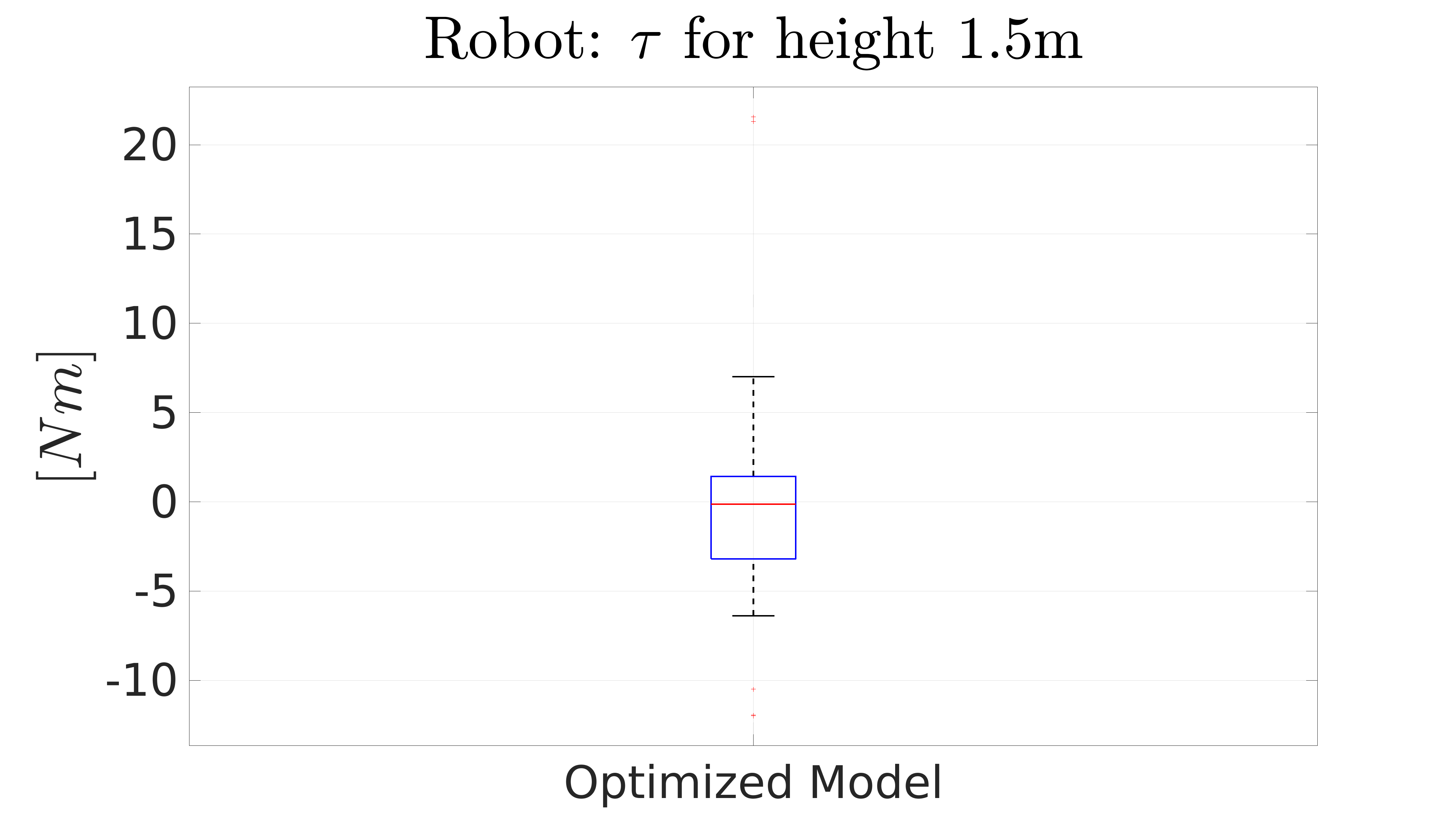}
  \caption{}
  \label{fig:robotTorque_4}
\end{subfigure}
\caption{Box plot of optimized robot and the original robot joint torques for the load height of $0.8 \text{ } \meter$ in (a), $1.0 \text{ } \meter$ in (b), $1.2 \text{ } \meter$ in (c) and  $1.5 \text{ } \meter$ in (d). Note that in (d) only the optimized robot is considered since the original robot is not able to reach the load due to its height. }
\label{fig:robotTorque_5}
\end{figure*}
\section{VALIDATION}
\label{sec:validation}
The proposed hardware parametrization and design optimization process has been tested starting from the humanoid robotic platform iCub \cite{Natale2017}. To introduce the link parametrization of Section \ref{sec:optimumHardware:hardware-parametrization}, the iCub robot has been modeled with simple shapes, i.e. \textit{sphere}, \textit{cylinder} and \textit{box}, resulting in the model showed in Fig. \ref{fig:all_results}. The shapes geometry can be described by the following variables: radius ($r$) for the sphere, width ($w$), depth ($d$), and height ($h$) for the box, height ($h$) and radius ($r$) for the cylinder. Inertial characteristics of each link have been parametrized with respect to $\left[ \rho, l_m \right]$. The link density $\rho$, is assumed to be constant in the link volume, while $l_m \in \mathbb{R}^+$ is a length multiplier that scales the shape geometry along the principal direction of the kinematic chain, hence, the direction in which the chain grows. The parameterization of the inertial characteristics of the possible link shapes can be found in Table \ref{tab:linkParam}. It is worth noticing that the inertia matrix has been computed with respect to a frame attached to the center of mass and that the principal directions are assumed to be $r$ for the radius, $d$ for the box, and $h$ for the cylinder.
The parametrization and the optimization problem formulated in Section \ref{sec:hardware-parametrization} have been defined using CasADi \cite{Andersson2019}, and the interior point method \cite{wachter2006implementation} has been used to solve the non-linear problem defined.
\begin{figure*}[!t]
\begin{subfigure}[b]{1.0\columnwidth}
  \includegraphics[trim=0.0cm 0.0cm 0.0cm 0.0cm, clip=true,width=1.0\columnwidth]{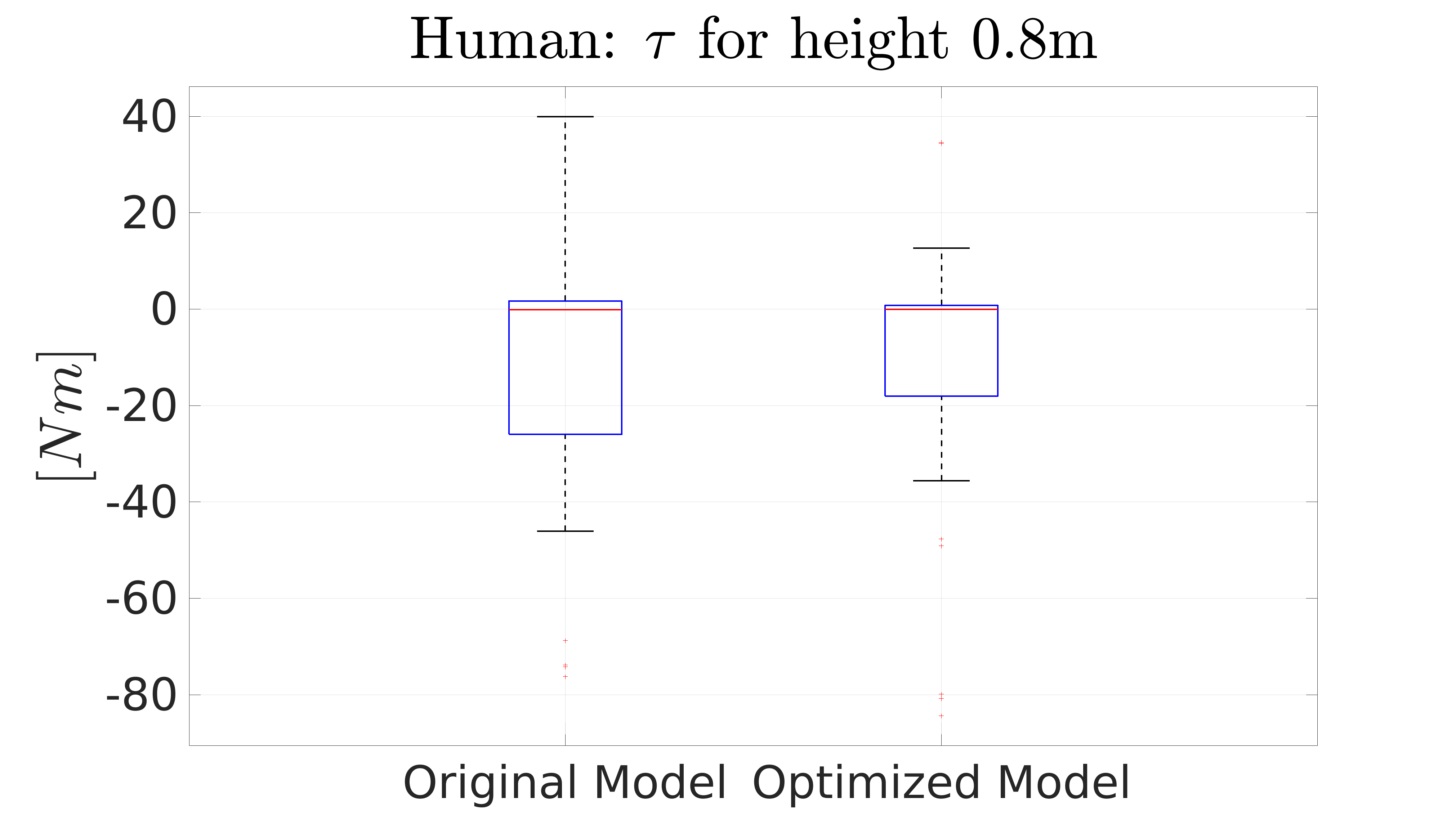}
  \caption{}
  \label{fig:humanTorque_1}
\end{subfigure}
\begin{subfigure}[b]{1.0\columnwidth}
  \includegraphics[trim=0.0cm 0.0cm 0.0cm 0.0cm, clip=true,width=1.0\columnwidth]{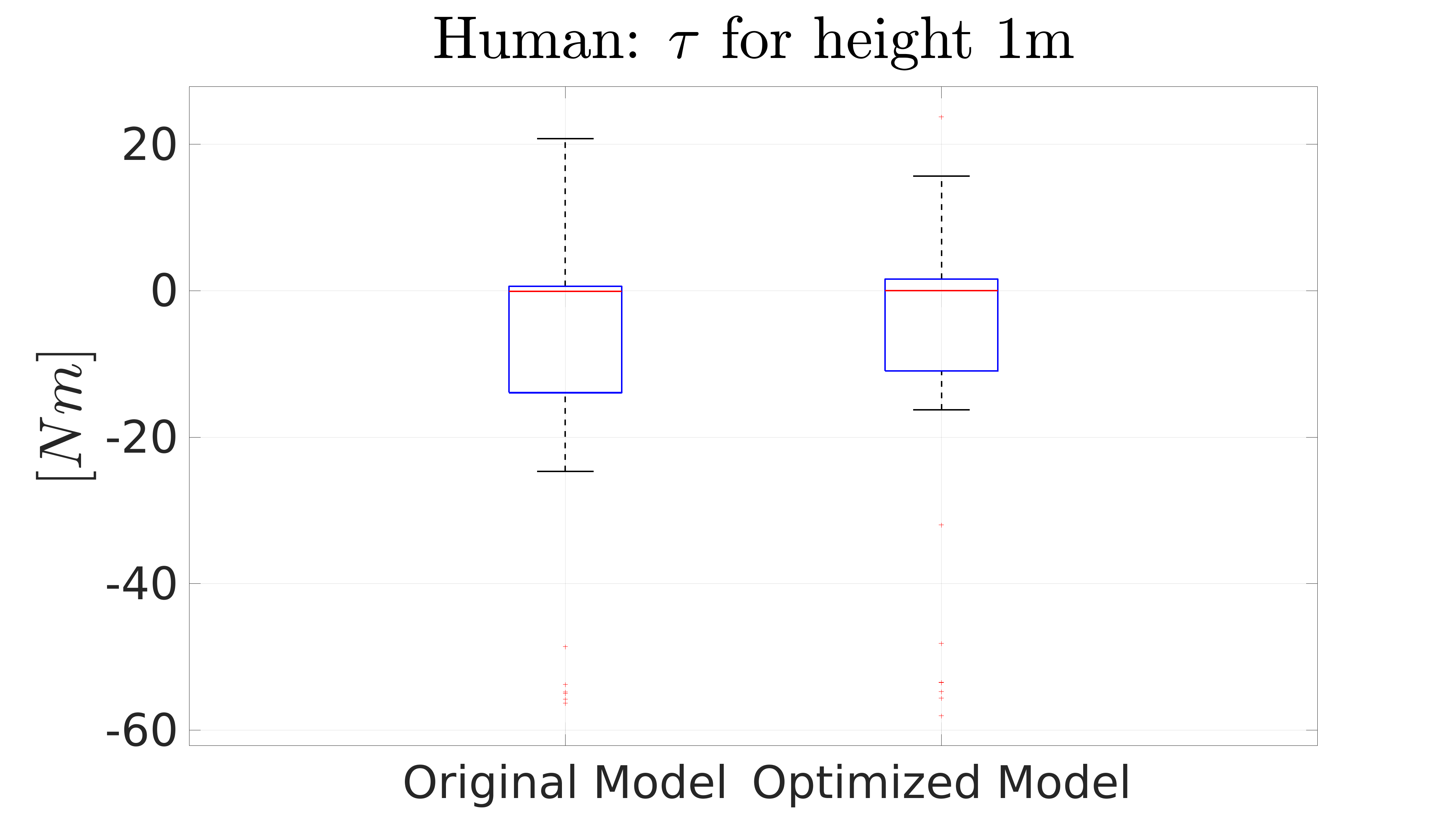}
  \caption{}
  \label{fig:humanTorque_2}
\end{subfigure}
\begin{subfigure}[b]{1.0\columnwidth}
  \includegraphics[trim=0.0cm 0.0cm 0.0cm 0.0cm, clip=true,width=1.0\columnwidth]{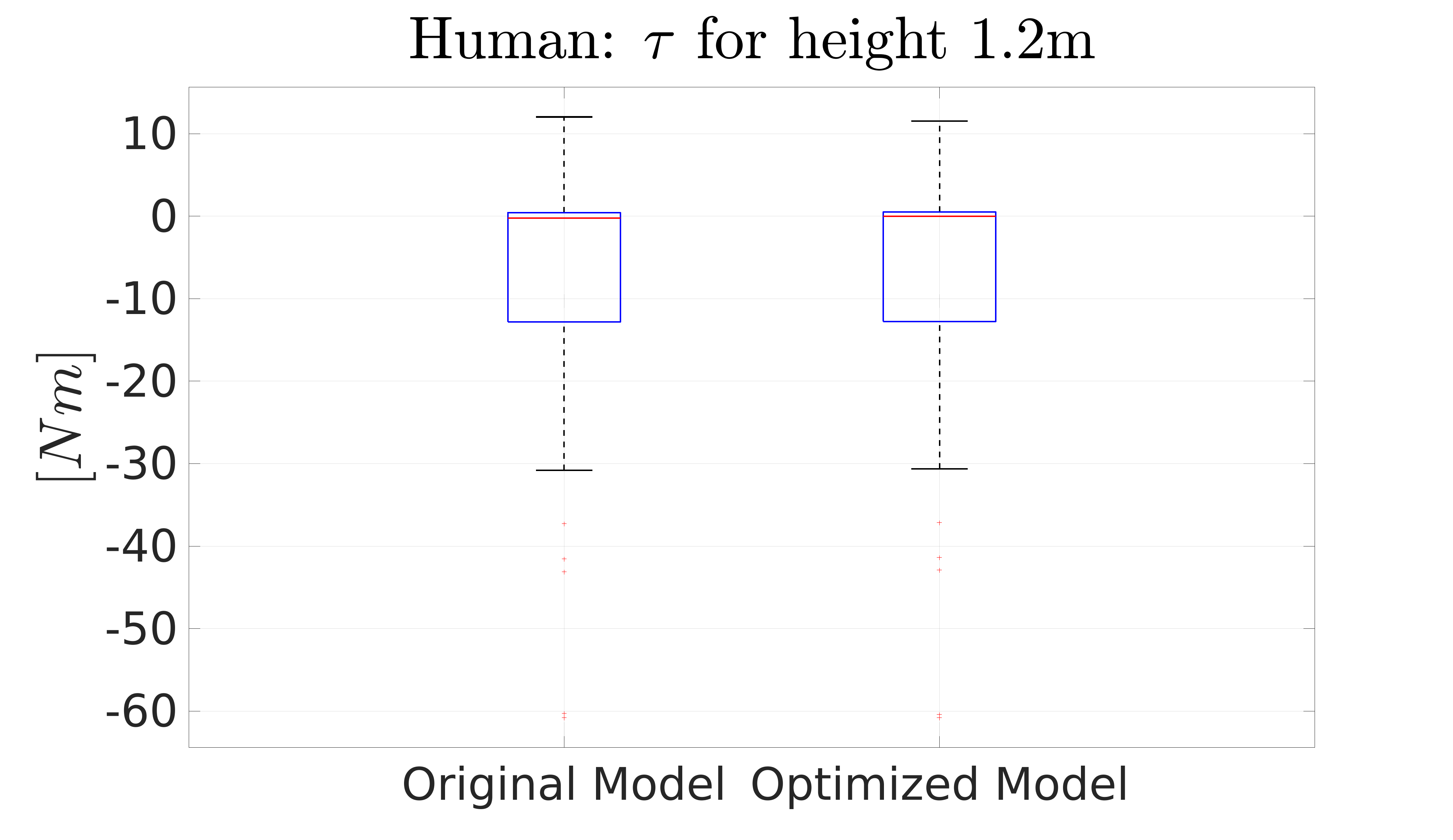}
  \caption{}
  \label{fig:humanTorque_3}
\end{subfigure}
\begin{subfigure}[b]{1.0\columnwidth}
  \includegraphics[trim=0.0cm 0.0cm 0.0cm 0.0cm, clip=true,width=1.0\columnwidth]{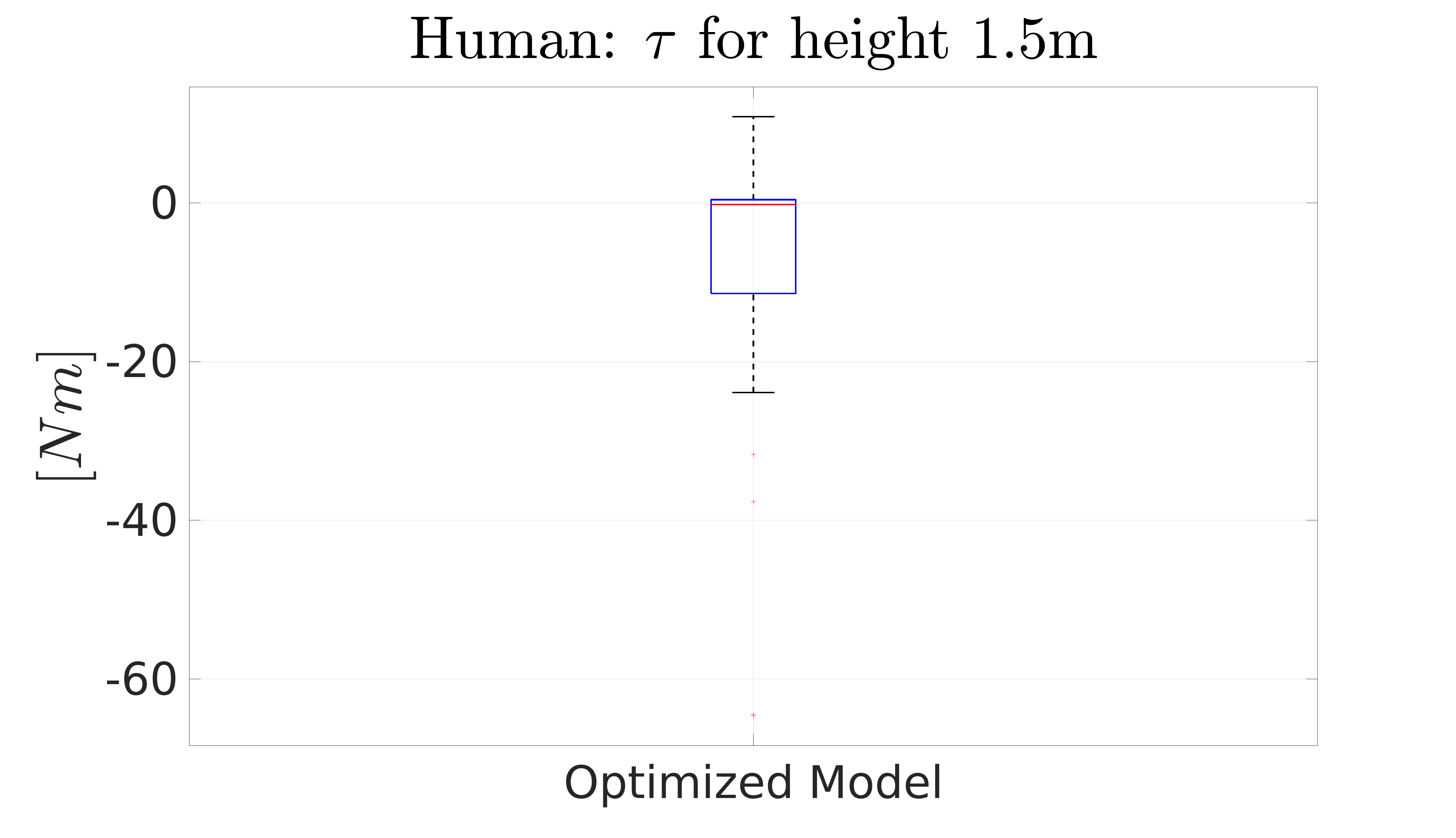}
  \caption{}
  \label{fig:humanTorque_4}
\end{subfigure}
\caption{Box plot of the human joint torques when collaborating with the optimized robot and the original robot for the load height of $0.8 \text{ } \meter$ in (a), $1.0 \text{ } \meter$ in (b), $1.2 \text{ } \meter$ in (c) and  $1.5 \text{ } \meter$ in (d). Note that in (d) only the optimized robot is considered since the original model is not able to reach the load due to its height. }
\label{fig:humanTorque}
\end{figure*}

\begin{table*}[!t]
\begin{center}
\caption{Torque norm  $\left\rVert\tau\right\rVert_2 \textit{ } \left[Nm\right ]$ for each payload height, for the original and the optimized robot.}
\label{tab:torques}
\begin{tabular}{|c|c|c|c|c|c|}
\hline
Model &Agent& \textbf{$0.8 \textit{ } \meter$ } & \textbf{$1.0 \textit{ } \meter$} & \textbf{$1.2 \textit{ } \meter$} & \textbf{$1.5 \textit{ } \meter$}\\ \hline
Original Model &Human &$181.7120$ & $140.6781$ & $120.8854$ &  -- \\\hline
Original Model & Robot &$41.0307$ & $144.2744$ & $20.9576$ & -- \\\hline
Optimized Model &Human & $192.4057$ & $141.5792$ & $120.5903$ & $113.0994$ \\\hline
Optimized Model & Robot & $58.7634$ &$90.4088$ &$23.5889$ & $38.7279$\\ \hline
\end{tabular}
\end{center}
\end{table*}

\subsection{Optimization Results}
\label{sec:validation:multi-agent}
The optimization problem of Eq. \eqref{eq:optimizationProblem} has been applied to identify humanoid robot hardware parameters to perform lifting tasks of a payload, placed at different heights, in collaboration with a human. For such an optimization, the set of considered hardware parameters was composed of the length multipliers and the densities for the torso, arms and legs, i.e. $\pi = [l_{m_1},..., l_{m_{n_l}}, \rho_{1},..., \rho_{{n_l}}]$. The human being has been modeled as a 48 DoF multi-body system 1.82 m tall \cite{latella2019simultaneous} and the payload has been considered as a box of dimensions $0.5 \text{ }\meter \times 0.5 \text{ }\meter \times 0.025 \text{ }\meter$ and of 5 {\kilogram}  weight. The obtained humanoid robot design is depicted in Fig. \ref{fig:all_results}. The optimization output, together with the wrenches exchanged by the agent to the ground, are depicted in Fig. \ref{fig:optimized} for the optimized model and Fig. \ref{fig:original} for the original one. The payload target heights considered are $0.8 \text{ } \meter$ (Fig. \ref{fig:optimized_0}, \ref{fig:original_0}), $1.0 \text{ }\meter$ (Fig. \ref{fig:optimized_1}, \ref{fig:original_1}), $1.2 \text{ } \meter$ (Fig. \ref{fig:optimized_2}, \ref{fig:original_2}) and $1.5 \text{ } \meter$ (Fig. \ref{fig:optimized_3}).  Note that for the last height considered, the output for the original model is missing, the reason is that the original model was not able to achieve such target, due to its height.
The mean and  variance, depicted as box plots, of the joint torques of the optimized model are compared to the one of the original model for the height of  $0.8 \text{ } \meter$ (Fig. \ref{fig:robotTorque_1}), $1.0 \text{ }\meter$ (Fig. \ref{fig:robotTorque_2}), $1.2 \text{ } \meter$ (Fig. \ref{fig:robotTorque_3}), while for the height of $1.5 \text{ } \meter$ only the optimized model one are depicted in Fig. \ref{fig:robotTorque_4}. 
As can be noticed, the optimized model shows a torque mean which is either smaller or equal to the one of the original model. 
The mean and variance, depicted as box plot, of the joint torques of the human both with the optimized model and the original one, can be found in   $0.8 \text{ } \meter$ (Fig. \ref{fig:humanTorque_1}), $1.0 \text{ }\meter$ (Fig. \ref{fig:humanTorque_2}), $1.2 \text{ } \meter$ (Fig. \ref{fig:humanTorque_3}), while for the height of $1.5 \text{ } \meter$ only the one with the optimized model are depicted in Fig. \ref{fig:humanTorque_4}. The torque mean of the human when collaborating with the optimized model is comparable to the one of the human collaborating with the original model, hence the human ergonomy is preserved even though the range of height is increased. 
The human and robot $\left\rVert \tau\right\rVert_2$, at each payload height, are depicted in Table \ref{tab:torques} for both the original and the optimized model. 
In the table, it can be noticed as the optimized model torques are generally comparable to the original one and, for the height of $1.0 \text{ } \meter$ the optimized model torque decreases by about $33\%$ w.r.t. the original model torque, while the human one is comparable in the two collaborations. 
From the plot and the table previously mentioned, it can be appreciated as the robot energy expenditure decreases with the optimized model meanwhile the human one is preserved, even though the range of height is increased from $0.8-1.2 \text{ } \meter$ to $0.8-1.5 \text{ } \meter$, improving the interaction. 

\section{CONLCUSIONS}
\label{sec:conclusion}
This paper presents a methodology to identify humanoid robot hardware optimized for specific task execution, namely collaborative payload lifting. The proposed strategy provides a physically-consistent parametrization with respect to links geometry and density which is used to identify optimum hardware parameters to design humanoid robots for ergonomic human-robot interaction. The presented results show that the obtained humanoid robot is able to reach a larger range of heights while decreasing the robot's energy expenditure and preserving the human one. 
The output of this work will be used in the context of the ergoCub project, to design the ergoCub humanoid robot, aimed at ergonomic physical collaboration with humans. 

In future work we plan to extend the proposed approach to consider also the time evolution of the system and to introduce constraints on the object manipulation, as well as constraints on the robot torques and wrenches, by introducing a proper model of the robot motors and complexifying the interaction model. In addition, we plan to evaluate the improvements in the human-robot interaction on the real ergoCub robotic platform. As final remarks, the tackled problem is characterized by several local minima, and the gradient-based method used showed limits, resulting in solutions influenced by the problem's initial conditions. For this reason, in the near future, we plan to move towards evolutionary algorithms to design the next ergoCub robots for ergonomic human robot interaction and to further improve the robot hardware, by considering not only different load heights but also different human models.  

\addtolength{\textheight}{-9.0cm}  






\bibliographystyle{IEEEtran}
\bibliography{bibliography}
\end{document}